\documentclass{sigchi}

\toappear{Permission to make digital or hard copies of all or part of this work for personal or classroom use is granted without fee provided that copies are not made or distributed for profit or commercial advantage and that copies bear this notice and the full citation on the first page. Copyrights for components of this work owned by others than ACM must be honored. Abstracting with credit is permitted. To copy otherwise, or republish, to post on servers or to redistribute to lists, requires prior specific permission and/or a fee. Request permissions from permissions@acm.org. \\
{\emph{L@S'15}}, March 14--March 15, 2015, Vancouver, Canada. \\
 Copyright \copyright~2015 ACM ISBN/15/03...\$15.00. \\}
\usepackage{comment}
\usepackage{balance}  % to better equalize the last page
\usepackage{times}    % comment if you want LaTeX's default font
\usepackage{url}      % llt: nicely formatted URLs
\usepackage{amsmath,epsfig} %citesort removed
\usepackage{amsfonts}
\usepackage{xcolor}
\usepackage{paralist}
\usepackage{booktabs}
\usepackage{graphicx}
\usepackage{caption}
\usepackage{subcaption}
\usepackage{color}
\usepackage{url}
\allowdisplaybreaks
\definecolor{dgreen}{rgb}{0,0.6,0}

\def\sams{\mathsf{MLP}}
\def\samss{\mathsf{MLP\text{-}S}}
\def\samsb{\mathsf{MLP\text{-}B}}

\usepackage{amssymb}
\usepackage{amsfonts}
\usepackage{mathrsfs}
\usepackage{xspace}
\usepackage{bm}
\usepackage{upgreek}

\newcommand{\safemath}[2]{\newcommand{#1}{\ensuremath{#2}\xspace}}

\safemath{\bma}{\mathbf{a}}
\safemath{\bmb}{\mathbf{b}}
\safemath{\bmc}{\mathbf{c}}
\safemath{\bmd}{\mathbf{d}}
\safemath{\bme}{\mathbf{e}}
\safemath{\bmf}{\mathbf{f}}
\safemath{\bmg}{\mathbf{g}}
\safemath{\bmh}{\mathbf{h}}
\safemath{\bmi}{\mathbf{i}}
\safemath{\bmj}{\mathbf{j}}
\safemath{\bmk}{\mathbf{k}}
\safemath{\bml}{\mathbf{l}}
\safemath{\bmm}{\mathbf{m}}
\safemath{\bmn}{\mathbf{n}}
\safemath{\bmo}{\mathbf{o}}
\safemath{\bmp}{\mathbf{p}}
\safemath{\bmq}{\mathbf{q}}
\safemath{\bmr}{\mathbf{r}}
\safemath{\bms}{\mathbf{s}}
\safemath{\bmt}{\mathbf{t}}
\safemath{\bmu}{\mathbf{u}}
\safemath{\bmv}{\mathbf{v}}
\safemath{\bmw}{\mathbf{w}}
\safemath{\bmx}{\mathbf{x}}
\safemath{\bmy}{\mathbf{y}}
\safemath{\bmz}{\mathbf{z}}
\safemath{\bmzero}{\mathbf{0}}
\safemath{\bmone}{\mathbf{1}}

\bmdefine{\biad}{a}
\bmdefine{\bibd}{b}
\bmdefine{\bicd}{c}
\bmdefine{\bidd}{d}
\bmdefine{\bied}{e}
\bmdefine{\bifd}{f}
\bmdefine{\bigd}{g}
\bmdefine{\bihd}{h}
\bmdefine{\biid}{i}
\bmdefine{\bijd}{j}
\bmdefine{\bikd}{k}
\bmdefine{\bild}{l}
\bmdefine{\bimd}{m}
\bmdefine{\bind}{n}
\bmdefine{\biod}{o}
\bmdefine{\bipd}{p}
\bmdefine{\biqd}{q}
\bmdefine{\bird}{r}
\bmdefine{\bisd}{s}
\bmdefine{\bitd}{t}
\bmdefine{\biud}{u}
\bmdefine{\bivd}{v}
\bmdefine{\biwd}{w}
\bmdefine{\bixd}{x}
\bmdefine{\biyd}{y}
\bmdefine{\bizd}{z}

\bmdefine{\bixid}{\xi}
\bmdefine{\bilambdad}{\lambda}
\bmdefine{\bimud}{\mu}
\bmdefine{\bithetad}{\theta}
\bmdefine{\biphid}{\phi}
\bmdefine{\bideltad}{\delta}

\safemath{\bmia}{\biad}
\safemath{\bmib}{\bibd}
\safemath{\bmic}{\bicd}
\safemath{\bmid}{\bidd}
\safemath{\bmie}{\bied}
\safemath{\bmif}{\bifd}
\safemath{\bmig}{\bigd}
\safemath{\bmih}{\bihd}
\safemath{\bmii}{\biid}
\safemath{\bmij}{\bijd}
\safemath{\bmik}{\bikd}
\safemath{\bmil}{\bild}
\safemath{\bmim}{\bimd}
\safemath{\bmin}{\bind}
\safemath{\bmio}{\biod}
\safemath{\bmip}{\bipd}
\safemath{\bmiq}{\biqd}
\safemath{\bmir}{\bird}
\safemath{\bmis}{\bisd}
\safemath{\bmit}{\bitd}
\safemath{\bmiu}{\biud}
\safemath{\bmiv}{\bivd}
\safemath{\bmiw}{\biwd}
\safemath{\bmix}{\bixd}
\safemath{\bmiy}{\biyd}
\safemath{\bmiz}{\bizd}

\safemath{\bmxi}{\bixid}
\safemath{\bmlambda}{\bilambdad}
\safemath{\bmmu}{\bimud}
\safemath{\bmtheta}{\bithetad}
\safemath{\bmphi}{\biphid}
\safemath{\bmdelta}{\bideltad}

\safemath{\bA}{\mathbf{A}}
\safemath{\bB}{\mathbf{B}}
\safemath{\bC}{\mathbf{C}}
\safemath{\bD}{\mathbf{D}}
\safemath{\bE}{\mathbf{E}}
\safemath{\bF}{\mathbf{F}}
\safemath{\bG}{\mathbf{G}}
\safemath{\bH}{\mathbf{H}}
\safemath{\bI}{\mathbf{I}}
\safemath{\bJ}{\mathbf{J}}
\safemath{\bK}{\mathbf{K}}
\safemath{\bL}{\mathbf{L}}
\safemath{\bM}{\mathbf{M}}
\safemath{\bN}{\mathbf{N}}
\safemath{\bO}{\mathbf{O}}
\safemath{\bP}{\mathbf{P}}
\safemath{\bQ}{\mathbf{Q}}
\safemath{\bR}{\mathbf{R}}
\safemath{\bS}{\mathbf{S}}
\safemath{\bT}{\mathbf{T}}
\safemath{\bU}{\mathbf{U}}
\safemath{\bV}{\mathbf{V}}
\safemath{\bW}{\mathbf{W}}
\safemath{\bX}{\mathbf{X}}
\safemath{\bY}{\mathbf{Y}}
\safemath{\bZ}{\mathbf{Z}}

\safemath{\bZero}{\mathbf{0}}
\safemath{\bOne}{\mathbf{1}}
\safemath{\bDelta}{\mathbf{\Delta}}
\safemath{\bLambda}{\mathbf{\UpLambda}}
\safemath{\bPhi}{\mathbf{\Upphi}}
\safemath{\bSigma}{\mathbf{\Upsigma}}
\safemath{\bOmega}{\mathbf{\Upomega}}
\safemath{\bTheta}{\mathbf{\Uptheta}}

\bmdefine{\biAd}{A}
\bmdefine{\biBd}{B}
\bmdefine{\biCd}{C}
\bmdefine{\biDd}{D}
\bmdefine{\biEd}{E}
\bmdefine{\biFd}{F}
\bmdefine{\biGd}{G}
\bmdefine{\biHd}{H}
\bmdefine{\biId}{I}
\bmdefine{\biJd}{J}
\bmdefine{\biKd}{K}
\bmdefine{\biLd}{L}
\bmdefine{\biMd}{M}
\bmdefine{\biNd}{N}
\bmdefine{\biOd}{O}
\bmdefine{\biPd}{P}
\bmdefine{\biQd}{Q}
\bmdefine{\biRd}{R}
\bmdefine{\biSd}{S}
\bmdefine{\biTd}{T}
\bmdefine{\biUd}{U}
\bmdefine{\biVd}{V}
\bmdefine{\biWd}{W}
\bmdefine{\biXd}{X}
\bmdefine{\biYd}{Y}
\bmdefine{\biZd}{Z}

\bmdefine{\biDelta}{\Delta}
\bmdefine{\biLambda}{\Lambda}
\bmdefine{\biPhi}{\Phi}
\bmdefine{\biSigma}{\Sigma}
\bmdefine{\biOmega}{\Omega}
\bmdefine{\biTheta}{\Theta}

\safemath{\bimA}{\biAd}
\safemath{\bimB}{\biBd}
\safemath{\bimC}{\biCd}
\safemath{\bimD}{\biDd}
\safemath{\bimE}{\biEd}
\safemath{\bimF}{\biFd}
\safemath{\bimG}{\biGd}
\safemath{\bimH}{\biHd}
\safemath{\bimI}{\biId}
\safemath{\bimJ}{\biJd}
\safemath{\bimK}{\biKd}
\safemath{\bimL}{\biLd}
\safemath{\bimM}{\biMd}
\safemath{\bimN}{\biNd}
\safemath{\bimO}{\biOd}
\safemath{\bimP}{\biPd}
\safemath{\bimQ}{\biQd}
\safemath{\bimR}{\biRd}
\safemath{\bimS}{\biSd}
\safemath{\bimT}{\biTd}
\safemath{\bimU}{\biUd}
\safemath{\bimV}{\biVd}
\safemath{\bimW}{\biWd}
\safemath{\bimX}{\biXd}
\safemath{\bimY}{\biYd}
\safemath{\bimZ}{\biZd}

\safemath{\bimDelta}{\biDelta}
\safemath{\bimLambda}{\biLambda}
\safemath{\bimPhi}{\biPhi}
\safemath{\bimSigma}{\biSigma}
\safemath{\bimOmega}{\biOmega}
\safemath{\bimTheta}{\biTheta}

\safemath{\setA}{\mathcal{A}}
\safemath{\setB}{\mathcal{B}}
\safemath{\setC}{\mathcal{C}}
\safemath{\setD}{\mathcal{D}}
\safemath{\setE}{\mathcal{E}}
\safemath{\setF}{\mathcal{F}}
\safemath{\setG}{\mathcal{G}}
\safemath{\setH}{\mathcal{H}}
\safemath{\setI}{\mathcal{I}}
\safemath{\setJ}{\mathcal{J}}
\safemath{\setK}{\mathcal{K}}
\safemath{\setL}{\mathcal{L}}
\safemath{\setM}{\mathcal{M}}
\safemath{\setN}{\mathcal{N}}
\safemath{\setO}{\mathcal{O}}
\safemath{\setP}{\mathcal{P}}
\safemath{\setQ}{\mathcal{Q}}
\safemath{\setR}{\mathcal{R}}
\safemath{\setS}{\mathcal{S}}
\safemath{\setT}{\mathcal{T}}
\safemath{\setU}{\mathcal{U}}
\safemath{\setV}{\mathcal{V}}
\safemath{\setW}{\mathcal{W}}
\safemath{\setX}{\mathcal{X}}
\safemath{\setY}{\mathcal{Y}}
\safemath{\setZ}{\mathcal{Z}}
\safemath{\emptySet}{\varnothing}

\safemath{\colA}{\mathscr{A}}
\safemath{\colB}{\mathscr{B}}
\safemath{\colC}{\mathscr{C}}
\safemath{\colD}{\mathscr{D}}
\safemath{\colE}{\mathscr{E}}
\safemath{\colF}{\mathscr{F}}
\safemath{\colG}{\mathscr{G}}
\safemath{\colH}{\mathscr{H}}
\safemath{\colI}{\mathscr{I}}
\safemath{\colJ}{\mathscr{J}}
\safemath{\colK}{\mathscr{K}}
\safemath{\colL}{\mathscr{L}}
\safemath{\colM}{\mathscr{M}}
\safemath{\colN}{\mathscr{N}}
\safemath{\colO}{\mathscr{O}}
\safemath{\colP}{\mathscr{P}}
\safemath{\colQ}{\mathscr{Q}}
\safemath{\colR}{\mathscr{R}}
\safemath{\colS}{\mathscr{S}}
\safemath{\colT}{\mathscr{T}}
\safemath{\colU}{\mathscr{U}}
\safemath{\colV}{\mathscr{V}}
\safemath{\colW}{\mathscr{W}}
\safemath{\colX}{\mathscr{X}}
\safemath{\colY}{\mathscr{Y}}
\safemath{\colZ}{\mathscr{Z}}

\safemath{\opA}{\mathbb{A}}
\safemath{\opB}{\mathbb{B}}
\safemath{\opC}{\mathbb{C}}
\safemath{\opD}{\mathbb{D}}
\safemath{\opE}{\mathbb{E}}
\safemath{\opF}{\mathbb{F}}
\safemath{\opG}{\mathbb{G}}
\safemath{\opH}{\mathbb{H}}
\safemath{\opI}{\mathbb{I}}
\safemath{\opJ}{\mathbb{J}}
\safemath{\opK}{\mathbb{K}}
\safemath{\opL}{\mathbb{L}}
\safemath{\opM}{\mathbb{M}}
\safemath{\opN}{\mathbb{N}}
\safemath{\opO}{\mathbb{O}}
\safemath{\opP}{\mathbb{P}}
\safemath{\opQ}{\mathbb{Q}}
\safemath{\opR}{\mathbb{R}}
\safemath{\opS}{\mathbb{S}}
\safemath{\opT}{\mathbb{T}}
\safemath{\opU}{\mathbb{U}}
\safemath{\opV}{\mathbb{V}}
\safemath{\opW}{\mathbb{W}}
\safemath{\opX}{\mathbb{X}}
\safemath{\opY}{\mathbb{Y}}
\safemath{\opZ}{\mathbb{Z}}
\safemath{\opZero}{\mathbb{O}}
\safemath{\identityop}{\opI}

\safemath{\veca}{\bma}
\safemath{\vecb}{\bmb}
\safemath{\vecc}{\bmc}
\safemath{\vecd}{\bmd}
\safemath{\vece}{\bme}
\safemath{\vecf}{\bmf}
\safemath{\vecg}{\bmg}
\safemath{\vech}{\bmh}
\safemath{\veci}{\bmi}
\safemath{\vecj}{\bmj}
\safemath{\veck}{\bmk}
\safemath{\vecl}{\bml}
\safemath{\vecm}{\bmm}
\safemath{\vecn}{\bmn}
\safemath{\veco}{\bmo}
\safemath{\vecp}{\bmp}
\safemath{\vecq}{\bmq}
\safemath{\vecr}{\bmr}
\safemath{\vecs}{\bms}
\safemath{\vect}{\bmt}
\safemath{\vecu}{\bmu}
\safemath{\vecv}{\bmv}
\safemath{\vecw}{\bmw}
\safemath{\vecx}{\bmx}
\safemath{\vecy}{\bmy}
\safemath{\vecz}{\bmz}

\safemath{\veczero}{\bmzero}
\safemath{\vecone}{\bmone}
\safemath{\vecxi}{\bmxi}
\safemath{\veclambda}{\bmlambda}
\safemath{\vecmu}{\bmmu}
\safemath{\vectheta}{\bmtheta}
\safemath{\vecphi}{\bmphi}
\safemath{\vecdelta}{\bmdelta}

\safemath{\matA}{\bA}
\safemath{\matB}{\bB}
\safemath{\matC}{\bC}
\safemath{\matD}{\bD}
\safemath{\matE}{\bE}
\safemath{\matF}{\bF}
\safemath{\matG}{\bG}
\safemath{\matH}{\bH}
\safemath{\matI}{\bI}
\safemath{\matJ}{\bJ}
\safemath{\matK}{\bK}
\safemath{\matL}{\bL}
\safemath{\matM}{\bM}
\safemath{\matN}{\bN}
\safemath{\matO}{\bO}
\safemath{\matP}{\bP}
\safemath{\matQ}{\bQ}
\safemath{\matR}{\bR}
\safemath{\matS}{\bS}
\safemath{\matT}{\bT}
\safemath{\matU}{\bU}
\safemath{\matV}{\bV}
\safemath{\matW}{\bW}
\safemath{\matX}{\bX}
\safemath{\matY}{\bY}
\safemath{\matZ}{\bZ}
\safemath{\matzero}{\bmzero}

\safemath{\matDelta}{\bDelta}
\safemath{\matLambda}{\bLambda}
\safemath{\matPhi}{\bPhi}
\safemath{\matSigma}{\bSigma}
\safemath{\matOmega}{\bOmega}
\safemath{\matTheta}{\bTheta}

\safemath{\matidentity}{\matI}
\safemath{\matone}{\matO}

\safemath{\rnda}{A}
\safemath{\rndb}{B}
\safemath{\rndc}{C}
\safemath{\rndd}{D}
\safemath{\rnde}{E}
\safemath{\rndf}{F}
\safemath{\rndg}{G}
\safemath{\rndh}{H}
\safemath{\rndi}{I}
\safemath{\rndj}{J}
\safemath{\rndk}{K}
\safemath{\rndl}{L}
\safemath{\rndm}{M}
\safemath{\rndn}{N}
\safemath{\rndo}{O}
\safemath{\rndp}{P}
\safemath{\rndq}{Q}
\safemath{\rndr}{R}
\safemath{\rnds}{S}
\safemath{\rndt}{T}
\safemath{\rndu}{U}
\safemath{\rndv}{V}
\safemath{\rndw}{W}
\safemath{\rndx}{X}
\safemath{\rndy}{Y}
\safemath{\rndz}{Z}

\safemath{\rveca}{\bimA}
\safemath{\rvecb}{\bimB}
\safemath{\rvecc}{\bimC}
\safemath{\rvecd}{\bimD}
\safemath{\rvece}{\bimE}
\safemath{\rvecf}{\bimF}
\safemath{\rvecg}{\bimG}
\safemath{\rvech}{\bimH}
\safemath{\rveci}{\bimI}
\safemath{\rvecj}{\bimJ}
\safemath{\rveck}{\bimK}
\safemath{\rvecl}{\bimL}
\safemath{\rvecm}{\bimM}
\safemath{\rvecn}{\bimN}
\safemath{\rveco}{\bomO}
\safemath{\rvecp}{\bimP}
\safemath{\rvecq}{\bimQ}
\safemath{\rvecr}{\bimR}
\safemath{\rvecs}{\bimS}
\safemath{\rvect}{\bimT}
\safemath{\rvecu}{\bimU}
\safemath{\rvecv}{\bimV}
\safemath{\rvecw}{\bimW}
\safemath{\rvecx}{\bimX}
\safemath{\rvecy}{\bimY}
\safemath{\rvecz}{\bimZ}

\safemath{\rvecxi}{\bmxi}
\safemath{\rveclambda}{\bmlambda}
\safemath{\rvecmu}{\bmmu}
\safemath{\rvectheta}{\bmtheta}
\safemath{\rvecphi}{\bmphi}

\safemath{\rmatA}{\bimA}
\safemath{\rmatB}{\bimB}
\safemath{\rmatC}{\bimC}
\safemath{\rmatD}{\bimD}
\safemath{\rmatE}{\bimE}
\safemath{\rmatF}{\bimF}
\safemath{\rmatG}{\bimG}
\safemath{\rmatH}{\bimH}
\safemath{\rmatI}{\bimI}
\safemath{\rmatJ}{\bimJ}
\safemath{\rmatK}{\bimK}
\safemath{\rmatL}{\bimL}
\safemath{\rmatM}{\bimM}
\safemath{\rmatN}{\bimN}
\safemath{\rmatO}{\bimO}
\safemath{\rmatP}{\bimP}
\safemath{\rmatQ}{\bimQ}
\safemath{\rmatR}{\bimR}
\safemath{\rmatS}{\bimS}
\safemath{\rmatT}{\bimT}
\safemath{\rmatU}{\bimU}
\safemath{\rmatV}{\bimV}
\safemath{\rmatW}{\bimW}
\safemath{\rmatX}{\bimX}
\safemath{\rmatY}{\bimY}
\safemath{\rmatZ}{\bimZ}

\safemath{\rmatDelta}{\bimDelta}
\safemath{\rmatLambda}{\bimLambda}
\safemath{\rmatPhi}{\bimPhi}
\safemath{\rmatSigma}{\bimSigma}
\safemath{\rmatOmega}{\bimOmega}
\safemath{\rmatTheta}{\bimTheta}

\usepackage{amssymb}
\usepackage{amsfonts}
\usepackage{mathrsfs}
\usepackage{xspace}
\usepackage{bm}
\usepackage{fancyref}
\usepackage{textcomp}

\usepackage{multirow}
\usepackage{stmaryrd}

\newenvironment{textbmatrix}{	\setlength{\arraycolsep}{2.5pt}%
								\big[\begin{matrix}}{\end{matrix}\big]%
								\raisebox{0.08ex}{\vphantom{M}}}

\def\be{\begin{equation}}
\def\ee{\end{equation}}
\def\een{\nonumber \end{equation}}
\def\mat{\begin{bmatrix}}
\def\emat{\end{bmatrix}}
\def\btm{\begin{textbmatrix}}
\def\etm{\end{textbmatrix}}

\def\ba#1\ea{\begin{align}#1\end{align}}
\def\bas#1\eas{\begin{align*}#1\end{align*}}
\def\bs#1\es{\begin{split}#1\end{split}} 
\def\bg#1\eg{\begin{gather}#1\end{gather}}
\def\bml#1\eml{\begin{multline}#1\end{multline}}
\def\bi#1\ei{\begin{itemize}#1\end{itemize}}

\safemath{\dirac}{\delta}					% Dirac delta
\safemath{\krond}{\dirac}					% Kronecker delta
\safemath{\upto}{\uparrow}
\safemath{\downto}{\downarrow}
\safemath{\iu}{j}							% imaginary unit
\safemath{\ev}{\lambda}						% eigenvalue
\safemath{\hilseqspace}{l^{2}}				% Hilbert sequence space
\newcommand{\banachfunspace}[1]{\setL^{#1}}	% Banach function space
\safemath{\hilfunspace}{\banachfunspace{2}}	% Hilbert function space
\safemath{\SNR}{\text{\sc snr}} 				% signal to noise ratio
\safemath{\No}{N_0}							% noise spectral density
\safemath{\Es}{E_s}							% energy per symbol
\safemath{\Eb}{E_b}							% energy per bit
\safemath{\EbNo}{\frac{\Eb}{\No}}
\safemath{\EsNo}{\frac{\Es}{\No}}

\DeclareMathOperator{\CHop}{\ensuremath{\opH}} % channel operator
\safemath{\tvir}{\rndh_{\CHop}}				% time-varying impulse response
\safemath{\tvtf}{\rndl_{\CHop}}				% 	-''- transfer function
\safemath{\spf}{\rnds_{\CHop}}				% spreading function
\safemath{\bff}{H_{\CHop}}					% bi-freuqency function

\safemath{\ircf}{r_{h}}						% impulse response correlation fn.
\safemath{\tftvcf}{r_{s}}					% scattering function
\safemath{\tfcf}{r_{l}}						% time-frequency correlation fn.
\safemath{\bfcf}{r_{H}}						% bi-frequency correlation fn.

\safemath{\tcorr}{c_h}						% time-correlation function
\safemath{\scf}{c_{s}}						% spreading function
\safemath{\tfcorr}{c_{l}}					% transfer-function correlation
\safemath{\fcorr}{c_{H}}						% frequency-correlation function

\safemath{\mi}{I}							% mutual information
\safemath{\capacity}{C}						% capacity

\safemath{\normal}{\mathcal{N}}			% normal distribution
\safemath{\jpg}{\mathcal{CN}}			% jointly proper Gaussian
\safemath{\mchain}{\leftrightarrow}		% Markov chain
\safemath{\dB}{\,\mathrm{dB}}
\safemath{\dBm}{\,\mathrm{dBm}}
\safemath{\Hz}{\,\mathrm{Hz}}
\safemath{\kHz}{\,\mathrm{kHz}}
\safemath{\MHz}{\,\mathrm{MHz}}
\safemath{\GHz}{\,\mathrm{GHz}}
\safemath{\s}{\,\mathrm{s}}
\safemath{\ms}{\,\mathrm{ms}}
\safemath{\mus}{\,\mathrm{\text{\textmu}s}}
\safemath{\ns}{\,\mathrm{ns}}
\safemath{\ps}{\,\mathrm{ps}}
\safemath{\meter}{\,\mathrm{m}}
\safemath{\mm}{\,\mathrm{mm}}
\safemath{\cm}{\,\mathrm{cm}}
\safemath{\m}{\,\mathrm{m}}
\safemath{\W}{\,\mathrm{W}}
\safemath{\mW}{\, \mathrm{mW}}
\safemath{\J}{\,\mathrm{J}}
\safemath{\K}{\,\mathrm{K}}
\safemath{\bit}{\,\mathrm{bit}}
\safemath{\nat}{\,\mathrm{nat}}

\safemath{\define}{\triangleq}			% definition

\safemath{\equivalent}{\sim}
\safemath{\distas}{\sim}					% distributed according to
\safemath{\sdiff}{\Delta}				% symmetric set difference

\safemath{\reals}{\mathbb{R}}
\safemath{\positivereals}{\reals_{+}}
\safemath{\integers}{\mathbb{Z}}
\safemath{\posint}{\integers_{+}}
\safemath{\naturals}{\mathbb{N}}
\safemath{\posnaturals}{\naturals_{+}}
\safemath{\complexset}{\mathbb{C}}
\safemath{\rationals}{\mathbb{Q}}

\newcommand*{\fancyrefapplabelprefix}{app}		% Appendix
\newcommand*{\fancyrefthmlabelprefix}{thm}		% Theorem
\newcommand*{\fancyreflemlabelprefix}{lem}		% Lemma
\newcommand*{\fancyrefcorlabelprefix}{cor}		% Corollary
\newcommand*{\fancyrefdeflabelprefix}{def}		% Definition
\newcommand*{\fancyrefalglabelprefix}{alg}		% Definition
\newcommand*{\fancyrefproplabelprefix}{prop}		% Property
\newcommand*{\fancyrefexmpllabelprefix}{exmpl}
\newcommand*{\fancyreftbllabelprefix}{tbl}
\frefformat{vario}{\fancyrefseclabelprefix}{Section~#1}
\frefformat{vario}{\fancyrefthmlabelprefix}{Theorem~#1}
\frefformat{vario}{\fancyreflemlabelprefix}{Lemma~#1}
\frefformat{vario}{\fancyrefcorlabelprefix}{Corrolary~#1}
\frefformat{vario}{\fancyrefdeflabelprefix}{Definition~#1}
\frefformat{vario}{\fancyrefalglabelprefix}{Algorithm~#1}
\frefformat{vario}{\fancyreffiglabelprefix}{Figure~#1}
\frefformat{vario}{\fancyrefapplabelprefix}{Appendix~#1} 
\frefformat{vario}{\fancyrefeqlabelprefix}{(#1)}
\frefformat{vario}{\fancyrefproplabelprefix}{Proposition~#1}
\frefformat{vario}{\fancyrefexmpllabelprefix}{Example~#1}
\frefformat{vario}{\fancyreftbllabelprefix}{Table~#1}

\safemath{\dictab}{[\,\dicta\,\,\dictb\,]}

\safemath{\ysig}{\bmy}
\safemath{\ysighat}{\hat{\ysig}}
\safemath{\ysigdim}{M}
\safemath{\xsig}{\bmx}
\safemath{\xsigdim}{N}
\safemath{\nx}{n_x}
\safemath{\zsig}{\bmz}
\safemath{\zsigdim}{\ysigdim}
\safemath{\rsig}{\bmr}
\safemath{\Adict}{\bA}
\safemath{\Adicttilde}{\widetilde{\Adict}}
\safemath{\Adictdim}{\outputdim\times\xsigdim}
\safemath{\avec}{\bma}
\safemath{\avectilde}{\tilde{\avec}}
\safemath{\Bdict}{\bB}
\safemath{\Bdicttilde}{\widetilde{\Bdict}}
\safemath{\Cdict}{\bC}
\safemath{\cvec}{\bmc}
\safemath{\Ddict}{\bD}
\safemath{\Ddictdim}{\ysigdim\times\xsigdim}
\safemath{\dvec}{\bmd}
\safemath{\Ddicttilde}{\widetilde{\bD}}
\safemath{\Bonb}{\bB}
\safemath{\bvec}{\bmb}
\safemath{\Bonbdim}{\ysigdim\times\ysigdim}
\safemath{\noise}{\bmn}
\safemath{\noisedim}{\ysigim}
\safemath{\err}{\bme}
\safemath{\errdim}{\ysigdim}
\safemath{\errset}{\setE}
\safemath{\nerr}{n_e}
\safemath{\delop}{\bP_\errset}
\safemath{\delopc}{\bP_{{\errset}^c}}

\safemath{\cplxi}{\imath}
\safemath{\cplxj}{\jmath}
\safemath{\dict}{\matD}
\safemath{\inputdim}{N}		% number of columns of dictionary D
\safemath{\outputdim}{M}		%number of rows of dictionary D
\safemath{\sparsity}{S}	%sparsity
\safemath{\inputdimA}{{N_a}}	%total number of elements in dictionary A
\safemath{\inputdimB}{{N_b}}	%total number of elements in dictionary B
\safemath{\elemA}{{n_a}}	%number of elements chosen from dictionary A
\safemath{\elemB}{{n_b}}	%number of elements chosen from dictionary B
\safemath{\resA}{\matR_a}	%restriction map to elements of dictionary A
\safemath{\resB}{\matR_b}	%restriction map to elements of dictionary B
\safemath{\subD}{\matS} %subdictionary
\safemath{\subA}{\matS_a} %subdictionary part of A
\safemath{\subB}{\matS_b} %subdictionary part of B
\safemath{\dicta}{\matA} 	% first subdictionary
\safemath{\dictb}{\matB} 	% second subdictionary
\safemath{\hollowS}{H}
\safemath{\hollowA}{H_a}
\safemath{\hollowB}{H_b}
\safemath{\cross}{Z}
\safemath{\coh}{\mu_d}			% coherence dictionary
\safemath{\coha}{\mu_a}			% coherence first subdictionary
\safemath{\cohb}{\mu_b}			% coherence second subdictionary
\safemath{\mubs}{\nu}	%block sub-coherence
\safemath{\cohm}{\mu_m} %mutual coherence
\safemath{\dictset}{\setD}	% set of dictionaries
\safemath{\dictsetp}{\dictset(\coh,\coha,\cohb)}	% set of dictionaries parametrized
\safemath{\dictsetgen}{\dictset_\text{gen}}
\safemath{\dictsetgenp}{\dictsetgen(\coh)}
\safemath{\dictsetonb}{\dictset_\text{onb}}
\safemath{\dictsetonbp}{\dictsetonb(\coh)}

\safemath{\leftside}{U}
\safemath{\rightsideA}{R_a}
\safemath{\rightsideB}{R_b}

\safemath{\indexS}{\setI_S} %set of indices participating in sub-dictionary S

\safemath{\na}{n_a}			% cardinality of set of linearly independent columns of first dictionary
\safemath{\nb}{n_b}			% cardinality of set of linearly independent columns of second dictionary
\safemath{\coeffa}{p_i}	%coefficients for columns of A
\safemath{\coeffb}{q_j}	%coefficients for columns of B
\safemath{\seta}{\setP}		% set of linearly independent columns of A
\safemath{\setb}{\setQ}     % set of linearly independent columns of B
\safemath{\setw}{\setW}	%set of n largest elements of w
\safemath{\setz}{\setZ}	%set of L-n largest elements of z
\safemath{\cola}{\veca}		% generic element of the dictionary A
\safemath{\colb}{\vecb}		% generic element of the dictionary B
\safemath{\cold}{\vecd}		% generic element of the dictionary D
\safemath{\inputvec}{\vecx} 	%coefficient vector (input)
\safemath{\error}{\vece}	%error vector
\safemath{\noiseout}{\vecz} 	%noisy output vector
\safemath{\inputvecel}{x}
\safemath{\inputveca}{\vecx_a}
\safemath{\inputvecb}{\vecx_b}
\safemath{\outputvec}{\vecy}	%output of Dictionary
\safemath{\lambdamin}{\lambda_{\mathrm{min}}}
\safemath{\elltwo}{\ell_2}
\safemath{\ellone}{\ell_1}
\safemath{\ellzero}{\ell_0}
\safemath{\ellinf}{\ell_\infty}
\safemath{\licard}{Z(\coh,\coha,\cohb)}
\safemath{\xsol}{\hat{x}}
\safemath{\xbord}{x_b}		%Solution at the border
\safemath{\xstat}{x_s}		%Solution stationary in l0 prob
\safemath{\xstatLone}{\tilde{x}_s}
\safemath{\order}{\mathcal{O}} %order notation (big O)
\safemath{\scales}{\Theta} %scales as
\safemath{\ones}{\mathbf{1}} %all ones matrix
\safemath{\zeroes}{\mathbf{0}} %all zeroes matrix
\safemath{\thlone}{\kappa(\coh,\cohb)} %treshold l1 problem
\safemath{\constoneA}{\delta} %constant in l1 theorem to save space
\safemath{\constoneB}{\epsilon} %constant in l1 theorem to save space
\safemath{\nlarge}{L}				   %num large elements
\safemath{\sumlarge}{S_\nlarge}
\safemath{\maxlarger}{P_\nlarge}	   % maximum in Gribonval and Nielsen
\safemath{\Pzero}{\textrm{P0}}	
\safemath{\Pone}{\textrm{P1}}
\safemath{\vecfir}{\vecw}			 % \vecv element of the kernel of the dictionary, \vecv=[\vecfir \vecsec]
\safemath{\vecsec}{\vecz}
\safemath{\elvecfir}{w}              % element of vecfir
\safemath{\elvecsec}{z}				 % element of vecsec
\safemath{\nlargefir}{n}
\safemath{\normout}{\gamma}
\safemath{\auxfun}{h}
\safemath{\supp}{\textrm{supp}}%support

\safemath{\indexa}{\ell}
\safemath{\indexb}{r}
\safemath{\indexc}{i}
\safemath{\indexd}{j}

\safemath{\project}{P}%projector

\makeatletter
\def\url@leostyle{%
  \@ifundefined{selectfont}{\def\UrlFont{\sf}}{\def\UrlFont{\small\bf\ttfamily}}}
\makeatother
\def\pprw{8.5in}
\def\pprh{11in}

\begin{document}

%\title{Math Language Processing and how to lose your fantasy matchup with Booooooo!!! !! in the last moment thanks to fxxxing DJax}

\title{Mathematical Language Processing: \\[2mm]
Automatic Grading and Feedback \\[2mm]
for Open Response Mathematical Questions}

\numberofauthors{1} 
\author{
%
% 1st. author
\alignauthor
Andrew S.\ Lan, Divyanshu Vats, Andrew E.\ Waters, Richard G.\ Baraniuk \\
       \affaddr{Rice University}\\
       \affaddr{Houston, TX 77005}\\
       \email{\{mr.lan, dvats, waters, richb\}@sparfa.com}
}

\maketitle

\begin{abstract}
While computer and communication technologies have provided effective means to scale up many aspects of education, the submission and grading of assessments such as homework assignments and tests remains a weak link.
In this paper, we study the problem of automatically grading the kinds of {\em open response mathematical questions} that figure prominently in STEM (science, technology, engineering, and mathematics) courses.  
Our data-driven framework for {\em mathematical language processing} ($\sams$) leverages solution data from a large number of learners to evaluate the correctness of their solutions, assign partial-credit scores, and provide feedback to each learner on the likely locations of any errors.  
$\sams$ takes inspiration from the success of natural language processing for text data and comprises three main steps.  
First, we convert each solution to an open response mathematical question into a series of numerical {\em features}.
Second, we {\em cluster} the features from several solutions to uncover the structures of correct, partially correct, and incorrect solutions.
We develop two different clustering approaches, one that leverages generic clustering algorithms and one based on Bayesian nonparametrics.
Third,  we {\em automatically grade} the remaining (potentially large number of) solutions based on their assigned cluster and one instructor-provided grade per cluster.
As a bonus, we can track the cluster assignment of each step of a multistep solution and determine when it departs from a cluster of correct solutions, which enables us to indicate the {\em likely locations of errors} to learners.
We test and validate $\sams$ on real-world MOOC data to demonstrate how it can substantially reduce the human effort required in large-scale educational platforms.\end{abstract}


%\category{H.5.m.}{Information Interfaces and Presentation (e.g. HCI)}{Miscellaneous}

\section{Introduction}
\label{sec:intro}

Large-scale educational platforms have the capability to revolutionize education by providing inexpensive, high-quality learning opportunities for millions of learners worldwide.  Examples of such platforms include massive open online courses (MOOCs) \cite{pritchardmooc,courserawebsite,khan,edxwebsite,guolick,googlemooc}, intelligent tutoring systems \cite{woolf08}, computer-based homework and testing systems \cite{erater,sapling,tesr,webassign}, and personalized learning systems \cite{ost}.  While computer and communication technologies have provided effective means to scale up the number of learners viewing lectures (via streaming video), reading the textbook (via the web), interacting with simulations (via a graphical user interface), and engaging in discussions (via online forums), the submission and grading of assessments such as homework assignments and tests remains a weak link.

There is a pressing need to find new ways and means to automate two critical tasks that are typically handled by the instructor or course assistants in a small-scale course: ({\em i})~grading of assessments, including allotting partial credit for partially correct solutions, and ({\em ii}) providing individualized feedback to learners on the locations and types of their errors.  

Substantial progress has been made on automated grading and feedback systems in several restricted domains, including essay evaluation using natural language processing (NLP) \cite{erater,ease}, computer program evaluation \cite{codehint,gulwanifeedback2014,kkhint,singh2013automated,aspiringminds}, and mathematical proof verification \cite{germangrade,metamath,mizar}.  

In this paper, we study the problem of automatically grading the kinds of {\em open response mathematical questions} that figure prominently in STEM (science, technology, engineering, and mathematics) education.  
To the best of our knowledge, there exist no tools to automatically evaluate and allot partial-credit scores to the solutions of such questions.
As a result, large-scale education platforms have resorted either to oversimplified multiple choice input and binary grading schemes (correct/incorrect), which are known to convey less information about the learners' knowledge than open response questions \cite{ofvsmc}, or peer-grading schemes \cite{jhuangedm,joachims}, which shift the burden of grading from the course instructor to the learners.\footnote{While peer grading appears to have some pedagogical value for learners \cite{pglearning}, each learner typically needs to grade several solutions from other learners for each question they solve, in order to obtain an accurate grade estimate.}

\subsection{Main Contributions}

\begin{figure}[tp]
\vspace{-0.0cm}
\centering
\includegraphics[scale=0.35]{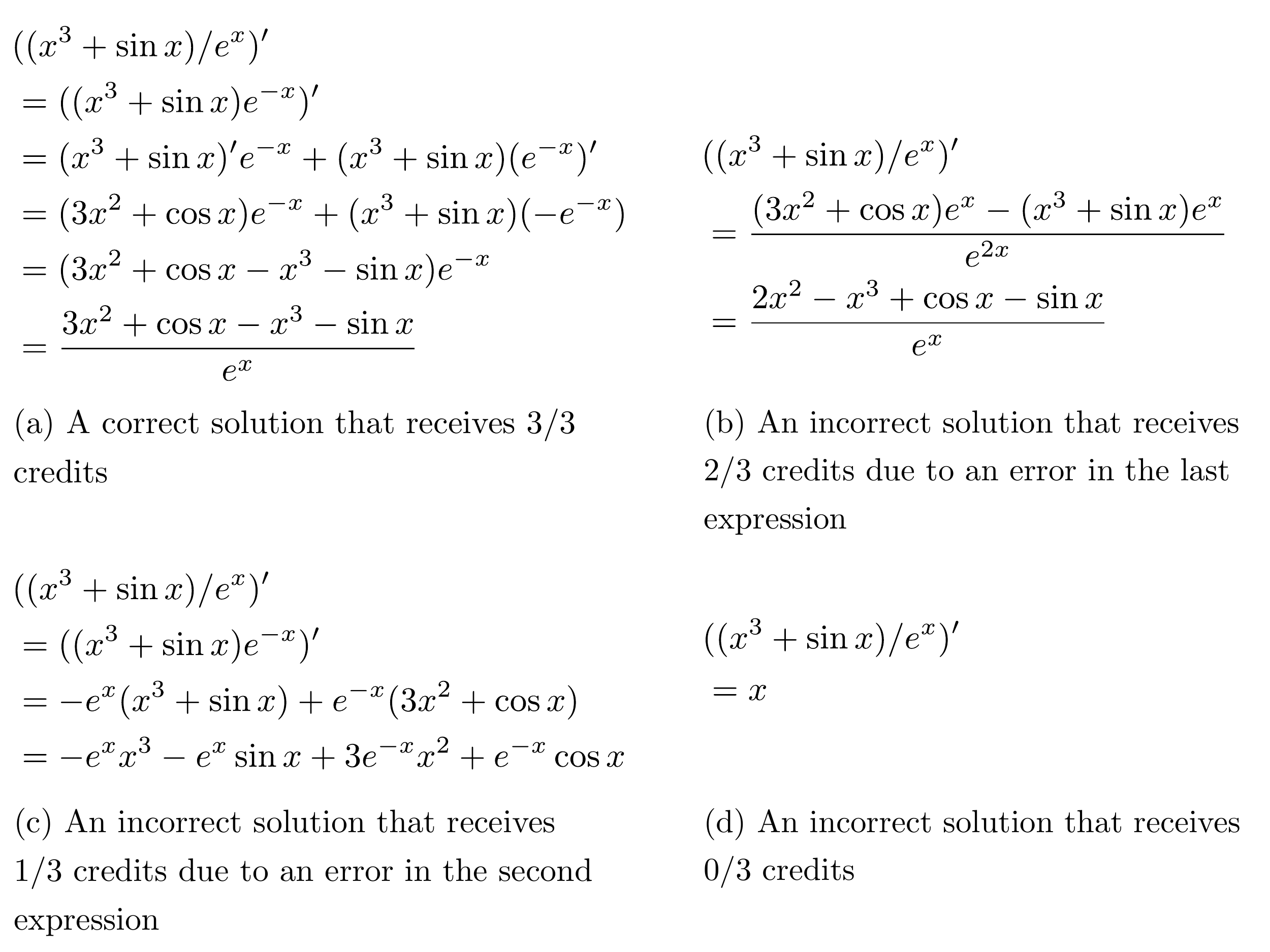}
\caption{Example solutions to the question ``Find the derivative of $(x^3+\sin x)/e^x$'' that were assigned scores of 3, 2, 1 and 0 out of 3, respectively, by our $\samsb$ algorithm.} 
\label{fig:ex1}
\end{figure}

\begin{figure}[t]
\vspace{-0.5cm}
\centering
\includegraphics[scale=0.36]{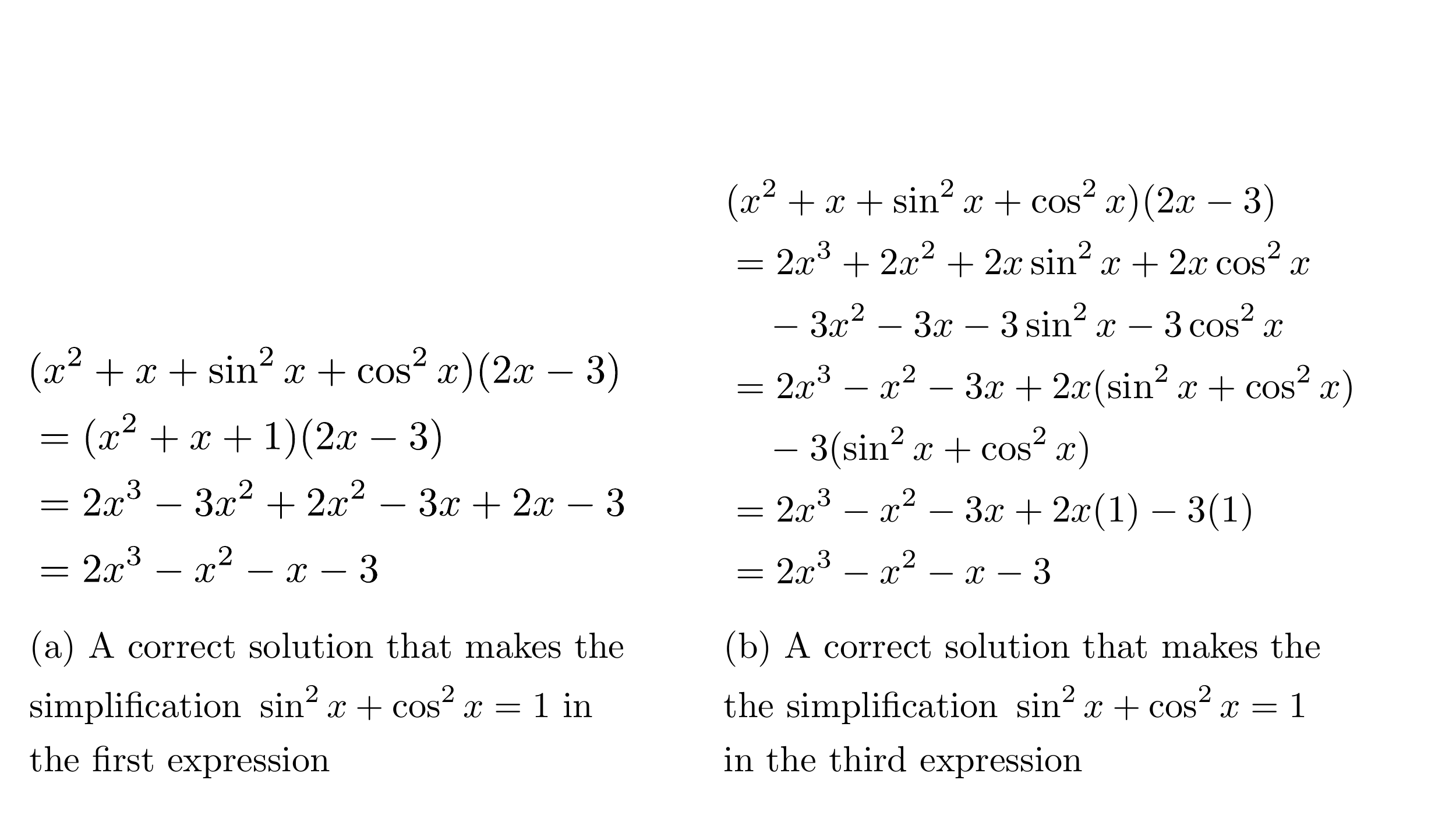}
\caption{Examples of two different yet correct paths to solve the question ``Simplify the expression $(x^2+x+\sin^2 x+ \cos^2 x) (2x-3) $.''} 
\label{fig:ex2}
\end{figure}

In this paper, we develop a data-driven framework for {\em mathematical language processing} ($\sams$) that leverages solution data from a large number of learners to evaluate the correctness of solutions to open response mathematical questions, assign partial-credit scores, and provide feedback to each learner on the likely locations of any errors.  
The scope of our framework is broad and covers questions whose solution involves one or more mathematical expressions.  
This includes not just formal proofs but also the kinds of mathematical calculations that figure prominently in science and engineering courses.  
Examples of solutions to two algebra questions of various levels of correctness are given in Figures~\ref{fig:ex1}~and~\ref{fig:ex2}.  
In this regard, our work differs significantly from that of \cite{germangrade}, which focuses exclusively on evaluating logical proofs.

Our $\sams$ framework, which is inspired by the success of NLP methods for the analysis of textual solutions (e.g., essays and short answer), comprises three main steps.  

First, we convert each solution to an open response mathematical question into a series of {\em numerical features}.  In deriving these features, we make use of symbolic mathematics to transform mathematical expressions into a canonical form.

Second, we {\em cluster} the features from several solutions to uncover the structures of correct, partially correct, and incorrect solutions.
We develop two different clustering approaches.
$\samss$ uses the numerical features to define a \textit{similarity score} between pairs of solutions and then applies a generic clustering algorithm, such as spectral clustering (SC) \cite{sc} or affinity propagation (AP) \cite{frey2007clustering}.  
We show that $\samss$ is also useful for visualizing mathematical solutions.  This can help instructors identify groups of learners that make similar errors so that instructors can deliver personalized remediation. 
$\samsb$ defines a {\em nonparametric Bayesian model} for the solutions and applies a Gibbs sampling algorithm to cluster the solutions.  
%$\samsb$ is more computationally intensive than $\samss$, but provides better auto-grading accuracy. 
%Furthermore, we use $\samsb$ to provide feedback to learners on expressions in their solutions where they have likely made an error, i.e., the expressions that make their solution incorrect.  

Third, once a human assigns a grade to at least one solution in each cluster, we automatically {\em grade} the remaining (potentially large number of) solutions based on their assigned cluster. 
As a bonus, in $\samsb$, we can track the cluster assignment of each step in a multistep solution and determine when it departs from a cluster of correct solutions, which enables us to indicate the likely locations of errors to learners.

In developing $\sams$, we tackle three main challenges of analyzing open response mathematical solutions.
%Several issues complicate making $\sams$ a reality.  
First, solutions might contain different notations that refer to the same mathematical quantity.  
For instance, in \fref{fig:ex1}, the learners use both $e^{-x}$ and $\frac{1}{e^x}$ to refer to the same quantity.  
Second, some questions admit more than one path to the correct/incorrect solution.  
For instance, in \fref{fig:ex2} we see two different yet correct solutions to the same question.  
It is typically infeasible for an instructor to enumerate all of these possibilities to automate the grading and feedback process.  
Third, numerically verifying the correctness of the solutions does not always apply to mathematical questions, especially when simplifications are required.
For example, a question that asks to simplify the expression $\sin^2 x + \cos^2 x + x$ can have both $1+x$ and $\sin^2 x + \cos^2 x + x$ as numerically correct answers, since both these expressions output the same value for all values of $x$.  However, the correct answer is $1+x$, since the question expects the learners to recognize that $\sin^2 x + \cos^2 x = 1$.
%For instance, in \fref{fig:ex2} the goal is to simplify a mathematical expression, and incorrect solutions could still be numerically correct.\footnote{For example, when solving the question in \fref{fig:ex2}, that asks to simplify the expression $x^2+x+\sin^2 x + \cos^2 x)(2x-3)$, both consider solving the question in \fref{fig:ex2}. If a learner submits the second expression in \fref{fig:ex2}(b) as their final solution, then the solution would still be numerically correct, but unsatisfactory since it is not in simplified form. {\bf rich: not clear}}
Thus, methods developed to check the correctness of computer programs and formulae by specifying a range of different inputs and checking for the correct outputs, e.g., \cite{singh2013automated}, cannot always be applied to accurately grade open response mathematical questions.  

\subsection{Related Work}
\label{sec:rw}

Prior work has led to a number of methods for grading and providing feedback to the solutions of certain kinds of open response questions.  
A linear regression-based approach has been developed to grade essays using features extracted from a training corpus using Natural Language Processing (NLP) \cite{erater,ease}.  
Unfortunately, such a simple regression-based model does not perform well when applied to the features extracted from mathematical solutions.  
%The authors in \cite{singh2013automated, gulwanifeedback2014}, propose several methods for grading and providing feedback on computer programs.  
Several methods have been developed for automated analysis of computer programs \cite{gulwanifeedback2014,singh2013automated}.  
However, these methods do not apply to the solutions to open response mathematical questions, since they lack the structure and compilability of computer programs.  
Several methods have also been developed to check the correctness of the logic in mathematical proofs \cite{germangrade,metamath,mizar}.
However, these methods apply only to mathematical proofs involving logical operations and not the kinds of open-ended mathematical calculations that are often involved in science and engineering courses.  

The idea of clustering solutions to open response questions into groups of similar solutions has been used in a number of previous endeavors: 
\cite{msearly,mslas} uses clustering to grade short, textual answers to simple questions; 
\cite{jhuangwww} uses clustering to visualize a large collection of computer programs; 
and
\cite{rivers2012canonicalizing} uses clustering to grade and provide feedback on computer programs. 
Although the high-level concept underlying these works is resonant with the $\sams$ framework, the feature building techniques used in $\sams$ are very different, since the structure of mathematical solutions differs significantly from short textual answers and computer programs. 

This paper is organized as follows.  
In the next section, we develop our approach to convert open response mathematical solutions to numerical features that can be processed by machine learning algorithms. 
We then develop $\samss$ and $\samsb$ and use real-world MOOC data to showcase their ability to accurately grade a large number of solutions based on the instructor's grades for only a small number of solutions, thus substantially reducing the human effort required in large-scale educational platforms. 
We close with a discussion and perspectives on future research directions.

\section{MLP Feature Extraction}
\label{sec:feature}

The first step in our $\sams$ framework is to transform a collection of solutions to an open response mathematical question into a set of numerical features.
In later sections, we show how the numerical features can be used to cluster and grade solutions as well as generate informative learner feedback.

A solution to an open response mathematical question will in general contain a mixture of explanatory text and core mathematical expressions.  
Since the correctness of a solution depends primarily on the mathematical expressions, we will ignore the text when deriving features.  
However, we recognize that the text is potentially very useful for automatically generating explanations for various mathematical expressions.  We leave this avenue for future work.

A workhorse of NLP is the \emph{bag-of-words} model; it has found tremendous success in text semantic analysis. %\cite{bleilda}.  
This model treats a text document as a collection of words and uses the frequencies of the words as numerical features to perform tasks like topic classification and document clustering \cite{bleilda,mslas}. 

A solution to an open response mathematical question consists of a series of {\em mathematical expressions} that are chained together by text, punctuation, or mathematical delimiters including $=$, $\leq$, $>$, $\propto$, $\approx$, etc.
For example, the solution in Figure~\ref{fig:ex1}(b) contains the expressions
$((x^3 + \sin x)/e^x)'$,
$((3x^2 + \cos x ) e^x - (x^3+\sin x)e^x))/e^{2x}$, and
$(2x^2 - x^3+\cos x-\sin x )/e^x$ that are all separated by the delimiter ``$=$".

$\sams$ identifies the unique mathematical expressions contained in the learners' solutions and uses them as features, effectively extending the bag-of-words model to use mathematical expressions as features rather than words.  
To coin a phrase, $\sams$ uses a novel {\em bag-of-expressions} model.

Once the mathematical expressions have been extracted from a solution, we parse them using SymPy, the open source Python library for symbolic mathematics \cite{sympy}.\footnote{In particular, we use the {\tt parse\_expr} function.}
SymPy has powerful capability for simplifying expressions.  
For example, $x^2 + x^2$ can be simplified to $2x^2$, and $e^x x^2 / e^{2x}$ can be simplifed to $e^{-x} x^2$.  
In this way, we can identify the equivalent terms in expressions that refer to the same mathematical quantity, resulting in more accurate features.  
In practice for some questions, however, it might be necessary to tone down the level of SymPy's simplification.  
For instance, the key to solving the question in Figure~\ref{fig:ex2} is to simplify the expression using the Pythagorean identity $\sin^2 x + \cos^2 x = 1$.
If SymPy is called on to perform such a simplification automatically, then it will not be possible to verify whether a learner has correctly navigated the simplification in their solution.  
For such problems, it is advisable to perform only arithmetic simplifications.  

After extracting the expressions from the solutions, we transform the expressions into numerical features.  
We assume that  $N$ learners submit solutions to a particular mathematical question.  
Extracting the expressions from each solution using SymPy yields a total of $V$ unique expressions across the $N$ solutions.

We encode the solutions in a integer-valued {\em solution feature matrix} $\bY \in \mathbb{N}^{V \times N}$ whose rows correspond to different expressions and whose columns correspond to different solutions; that is, the $(i,j)^\text{th}$ entry of $\bY$ is given by
\begin{align*}  
%\label{eq:y}
Y_{i,j} = \text{times expression $i$ appears in solution $j$.}
\end{align*} 
Each column of $\bY$ corresponds to a numerical representation of a mathematical solution.  
Note that we do not consider the ordering of the expressions in this model; such an extension is an interesting avenue for future work.
In this paper, we indicate in $\bY$ only the presence and not the frequency of an expression, i.e., $\bY \in \{0,1\}^{V \times N}$ and
\begin{align}  
\label{eq:y}
Y_{i,j} = \left \{ \begin{array}{ll}
1  &\text{if} \,\, \text{expression $i$ appears in solution $j$} \\[0.1cm]
0 & \text{otherwise.}
\end{array} \right. \,
\end{align} 
The extension to encoding frequencies is straightforward.  

To illustrate how the matrix $\bY$ is constructed, consider the solutions in \fref{fig:ex2}(a) and (b).  
Across both solutions, there are $7$ unique expressions.
Thus, $\bY$ is a $7 \times 2$ matrix, with each row corresponding to a unique expression.
Letting the first four rows of $\bY$ correspond to the four expressions in \fref{fig:ex2}(a) and the remaining three rows to expressions~2--4 in \fref{fig:ex2}(b), we have
\begin{align*}
\bY = \begin{bmatrix} 1 & 1 & 1 & 1 & 0 & 0 & 0 \\ 1 & 0 & 0 & 1 & 1 & 1 & 1 \end{bmatrix}^T.
\end{align*}

We end this section with the crucial observation that, for a wide range of mathematical questions, many expressions will be shared across learners' solutions.  
This is true, for instance, in Figure~\ref{fig:ex2}. 
This suggests that there are a limited number of types of solutions to a question (both correct and incorrect) and that solutions of the same type tend tend to be similar to each other.  
This leads us to the conclusion that the $N$ solutions to a particular question can be effectively clustered into $K \ll N$ clusters.  
In the next two sections, we will develop $\samss$ and $\samsb$, two algorithms to cluster solutions according to their numerical features.

\begin{figure*}[!t]
\vspace{-.0cm}
\begin{center}
\includegraphics[scale=.76]{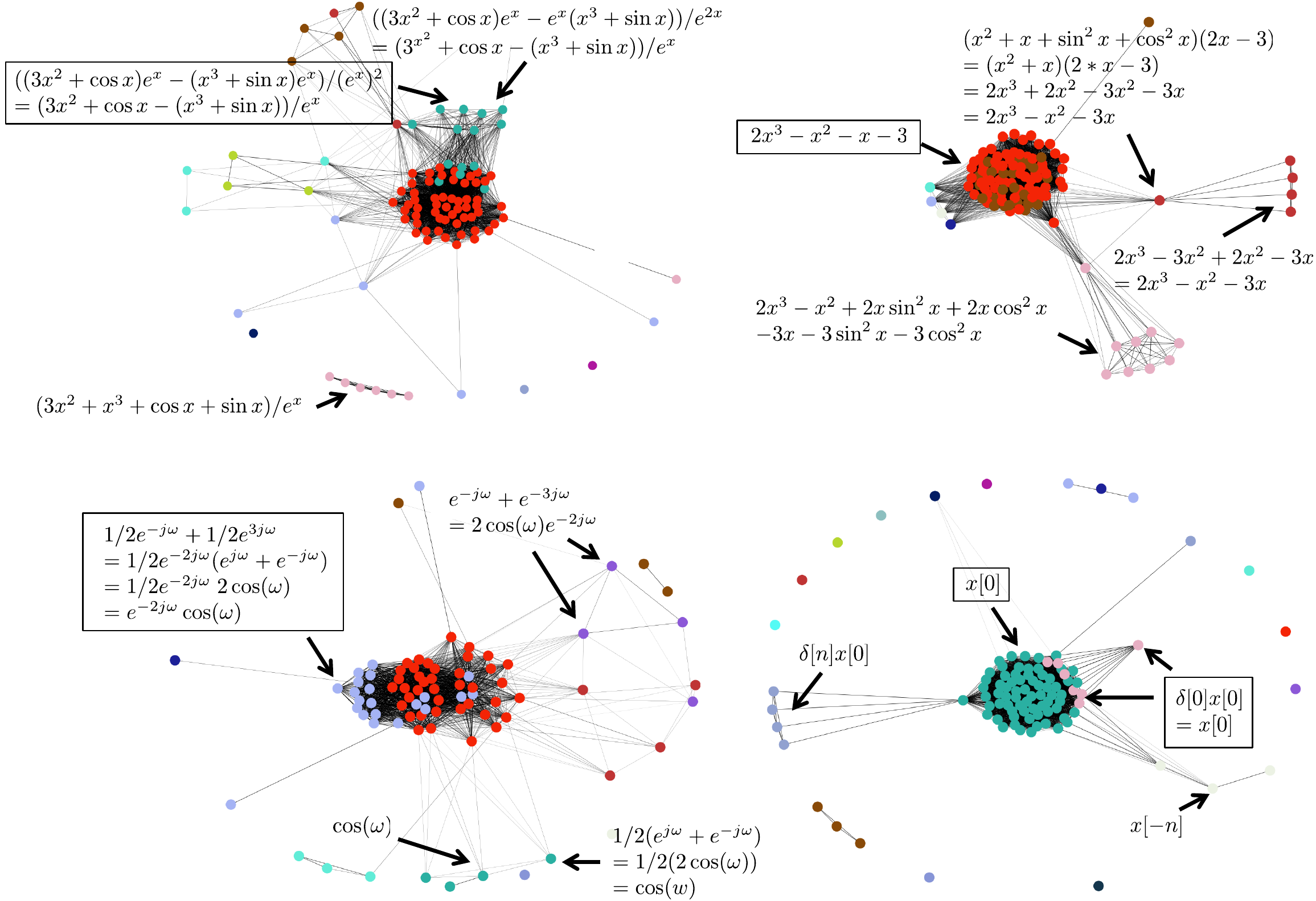}
\end{center}
\caption{Illustration of the clusters obtained by $\samss$ by applying affinity propagation (AP) on the similarity matrix $\bS$ corresponding to learners' solutions to four different mathematical questions (see Table~1 for more details about the datasets and the Appendix for the question statements).  Each node corresponds to a solution.  Nodes with the same color correspond to solutions that are estimated to be in the same cluster. The thickness of the edge between two solutions is proportional to their similarity score.  Boxed solutions are correct; all others are in varying degrees of correctness.
}
\label{fig:cluster}
\end{figure*}

\section{MLP-S: Similarity-Based Clustering}
\label{sec:jtree}

In this section, we outline $\samss$, which clusters and then grades solutions using a solution similarity-based approach.

\subsection{The $\samss$ Model}
\label{sec:metric}

We start by using the solution features in $\bY$ to define a notion of similarity between pairs of solutions.  
Define the $N \times N$ {\em similarity matrix} $\bS$ containing the pairwise similarities between all solutions, with its $(i,j)^\text{th}$ entry the similarity between solutions $i$ and $j$ 
\begin{equation}
\label{eq:similarity}
S_{i,j} = \frac{ \vecy_i^T \vecy_j}{\min\{\vecy_i^T \vecy_i, \vecy_j^T \vecy_j \}}\,.
\end{equation}
The column vector $\vecy_i$ denotes the $i^\text{th}$ column of $\bY$ and corresponds to learner $i$'s solution.
%It is clear that $S_{i,j}$ characterizes the similarity between solutions $\vecy_i$ and $\vecy_j$.  
Informally, $S_{i,j}$ is the number of common expressions between solution $i$ and solution $j$ divided by the minimum of the number of expressions in solutions $i$ and $j$.  
A large/small value of $S_{i,j}$ corresponds to the two solutions being similar/dissimilar.
For example, the similarity between the solutions in Figure~\ref{fig:ex1}(a) and Figure~\ref{fig:ex1}(b) is $1/3$ and the similarity between the solutions in Figure~\ref{fig:ex2}(a) and Figure~\ref{fig:ex2}(b) is $1/2$.
$\bS$ is symmetric, and $0 \le S_{i,j} \le 1$.  
Equation (\ref{eq:similarity}) is just one of any possible solution similarity metrics.
We defer the development of other metrics to future work. 

\subsection{Clustering Solutions in $\samss$}
\label{sec:cluster}

Having defined the similarity $S_{i,j}$ between two solutions $i$ and $j$, we now cluster the $N$ solutions into $K \ll N$ clusters such that the solutions within each cluster have high similarity score between them and solutions in different clusters have low similarity score between them.

Given the similarity matrix $\bS$, we can use any of the multitude of standard clustering algorithms to cluster solutions.  Two examples of clustering algorithms are {\em spectral clustering} (SC) \cite{sc} and {\em affinity propagation} (AP) \cite{frey2007clustering}.
The SC algorithm requires specifying the number of clusters $K$ as an input parameter, while the AP algorithm does not.%require specifying this parameter.

Figure~\ref{fig:cluster} illustrates how AP is able to identify clusters of similar solutions from solutions to four different mathematical questions.  The figures on the top correspond to solutions to the questions in Figures~\ref{fig:ex1} and~\ref{fig:ex2}, respectively.  The bottom two figures correspond to solutions to two signal processing questions.  Each node in the figure corresponds to a solution, and nodes with the same color correspond to solutions that belong to the same cluster.  For each figure, we show a sample solution from some of these clusters, with the boxed solutions corresponding to correct solutions.  We can make three interesting observations from \fref{fig:cluster}:

\begin{itemize}

\item In the top left figure, we cluster a solution with the final answer $3x^2 + \cos x - (x^3 + \sin x))/e^x$ with a solution with the final answer $3^{x^2} + \cos x - (x^3+\sin x))/e^x$.  Although the later solution is incorrect, it contained a typographical error where $3 * x \wedge 2$ was typed as $3 \wedge x \wedge 2$.  $\samss$ is able to identify this typographical error, since the expression before the final solution is contained in several other correct solutions.

\item In the top right figure, the correct solution requires identifying the trigonometric identify $\sin^2 x + \cos^2 x = 1$.  The clustering algorithm is able to identify a subset of the learners who were not able to identify this relationship and hence could not simplify their final expression.

\item $\samss$ is able to identify solutions that are strongly connected to each other.   Such a visualization can be extremely useful for course instructors.  For example, an instructor can easily identify a group of learners who lack mastery of a certain skill that results in a common error and adjust their course plan accordingly to help these learners. 

\end{itemize}

\subsection{Auto-Grading via $\samss$}

Having clustered all solutions into a small number $K$ of clusters, we assign the same grade to all solutions in the same cluster.
If a course instructor assigns a grade to one solution from each cluster, then $\samss$ can automatically grade the remaining $N-K$ solutions.  
We construct the index set $\mathcal{I}_S$ of solutions that the course instructor needs to grade as
\begin{align*}
\mathcal{I}_S = \left\{\arg \max_{i \in {\cal C}_k} \sum_{j = 1}^N S_{i,j}, \, k = 1,2,\ldots, K \right\},
\end{align*}
where ${\cal C}_k$ represents the index set of the solutions in cluster $k$.
%, and $S_{i,j}$ is the similarity defined in \fref{eq:similarity}. 
In words, in each cluster, we select the solution having the highest similarity to the other solutions (ties are broken randomly) to include in $\mathcal{I}_S$.  We demonstrate the performance of auto-grading via $\samss$ in the experimental results section below.

\section{MLP-B: Bayesian nonparametric clustering}
\label{sec:crp}

\fussy

%Recall again that we are given $N$ mathematical solutions and our aim is to grade all $N$ solutions on an ordinal scale.  
In this section, we outline $\samsb$, which clusters and then grades solutions using a Bayesian nonparameterics-based approach.  
%The steps involved in $\samsb$ are similar to those of $\samss$, with the key difference that $\samsb$ uses Bayesian nonparameterics to cluster and grade solutions.  
The $\samsb$ model and algorithm can be interpreted as an extension of the model in \cite{shorttext}, where a similar approach is proposed to cluster short text documents.

\subsection{The $\samsb$ Model}
\label{sec:model}

Following the key observation that the $N$ solutions can be effectively clustered into $K \ll N$ clusters, let $\vecz$ be the $N \times 1$ cluster assignment vector, with $z_j \in \{1,\ldots,K\}$ denoting the cluster assignment of the $j^\text{th}$ solution with $j \in \{1,\ldots,N\}$. Using this latent variable, we model the probability of the solution of all learners' solutions to the question as
\begin{align*}
p(\bY) = \prod_{j=1}^N \left( \sum_{k=1}^K p(\vecy_j | z_j = k) p(z_j = k) \right),
\end{align*}
where $\vecy_j$, the $j^\text{th}$ column of the data matrix $\bY$, corresponds to learner $j$'s solution to the question. Here we have implicitly assumed that the learners' solutions are independent of each other. 
By analogy to topic models \cite{bleilda,tm}, we assume that learner $j$'s solution to the question, $\vecy_j$, is generated according to a \emph{multinomial} distribution given the cluster assignments $\vecz$ as
 \begin{align} \label{eq:lik}
\notag p(\vecy_j | z_j = k) & = \textit{Mult}(\vecy_j | \boldsymbol{\phi}_k) \\ 
& = \frac{ (\sum_i Y_{i,j})!}{Y_{1,j}! Y_{2,j}! \ldots Y_{V,j}!} \Phi_{1,k}^{Y_{1,j}} \Phi_{2,k}^{Y_{2,j}} \ldots \Phi_{V,k}^{Y_{V,j}},
\end{align}
where $\boldsymbol{\Phi} \in [0,1]^{V \times K}$ is a parameter matrix with $\Phi_{v,k}$ denoting its $(v,k)^\text{th}$ entry. 
$\boldsymbol{\phi}_k \in [0,1]^{V \times 1}$ denotes the $k^\text{th}$ column of $\boldsymbol{\Phi}$ and charcterizes the multinomial distribution over all the $V$ features for cluster $k$. 

In practice, one often has no information regarding the number of clusters $K$. Therefore, we consider $K$ as an unknown parameter and infer it from the solution data. In order to do so, we impose a Chinese restaurant process (CRP) prior on the cluster assignments $\vecz$, parameterized by a parameter $\alpha$. The CRP characterizes the random partition of data into clusters, in analogy to the seating process of customers in a Chinese restaurant.
It is widely used in Bayesian mixture modeling literature \cite{crptm,crp}. Under the CRP prior, the cluster (table) assignment of the $j^\text{th}$ solution (customer), conditioned on the cluster assignments of all the other solutions, follows the distribution
\begin{align} \label{eq:crp}
p(z_j = k | \vecz_{\neg j}, \alpha) = \left \{ \begin{array}{ll}
\frac{n_{k,\neg j}}{N-1+\alpha}  & \text{if} \,\,\, \text{cluster $k$ is occupied}, \\[0.1cm]
\frac{\alpha}{N-1+\alpha} & \text{if} \,\,\, \text{cluster $k$ is empty},
\end{array} \right.
\end{align}
where $n_{k,\neg j}$ represents the number of solutions that belong to cluster $k$ excluding the current solution $j$, with $\sum_{k=1}^K n_{k,\neg j} = N-1$.  The vector $\vecz_{\neg j}$ represents the cluster assignments of the other solutions.  The flexibility of allowing any solution to start a new cluster of its own enables us to automatically infer $K$ from data. It is known \cite{tehdirichlet} that the expected number of clusters under the CRP prior satisfies $K \sim O(\alpha \, \text{log}N) \ll N$, so our method scales well as the number of learners $N$ grows large. We also impose a Gamma prior $\alpha \sim \textit{Gam}(\alpha_\alpha, \alpha_\beta)$ on $\alpha$ to help us infer its value. 

Since the solution feature data $\bY$ is assumed to follow a multinomial distribution parameterized by $\boldsymbol{\Phi}$, we impose a symmetric Dirichlet prior over $\boldsymbol{\Phi}$ as $\boldsymbol{\phi}_k \sim \textit{Dir} (\boldsymbol{\phi}_k | \beta)$ because of its conjugacy with the multinomial distribution \cite{gelmanbook}.

\begin{figure}[tp]
\vspace{-0.2cm}
\centering
\includegraphics[scale=0.19]{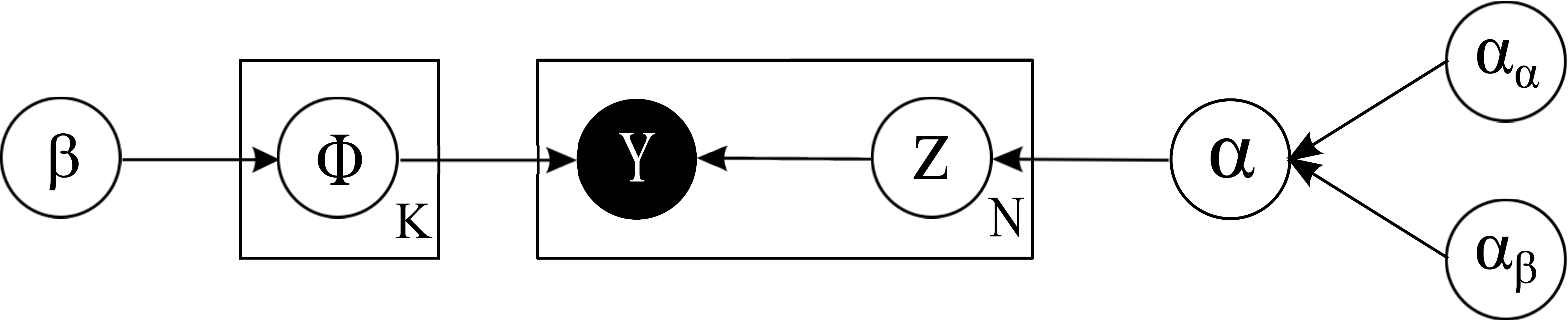}
\vspace{.0cm}
\caption{Graphical model of the generation process of solutions to mathematical questions. $\alpha_\alpha$, $\alpha_\beta$ and $\beta$ are hyperparameters, $\vecz$ and $\boldsymbol{\Phi}$ are latent variables to be inferred, and $\bY$ is the observed data defined in \fref{eq:y}.}
\label{fig:plate}
\vspace{-0.0cm}
\end{figure}

The graphical model representation of our model is visualized in \fref{fig:plate}. 
Our goal next is to estimate the cluster assignments $\vecz$ for the solution of each learner, the parameters $\boldsymbol{\phi}_k$ of each cluster, and the number of clusters $K$, from the binary-valued solution feature data matrix $\bY$. 

\subsection{Clustering Solutions in $\samsb$}
\label{sec:inference}

We use a Gibbs sampling algorithm for posterior inference under the $\samsb$ model, which automatically groups solutions into clusters.  We start by applying a generic clustering algorithm (e.g., $K$-means, with $K = N/10$) to initialize $\vecz$, and then initialize $\boldsymbol{\Phi}$ accordingly. Then, in each iteration of $\samsb$, we perform the following steps:
\begin{itemize}

\item[1.] {\bf Sample $\vecz$}: For each solution $j$, we remove it from its current cluster and sample its cluster assignment $z_j$ from the posterior $p(z_j = k | \vecz_{\neg j}, \alpha, \bY)$.  Using Bayes rule, we have 
\begin{align*}
  p(z_j = k | \vecz_{\neg j}, \boldsymbol{\Phi}, \alpha, \bY) 
& = p(z_j = k | \vecz_{\neg j}, \boldsymbol{\phi}_k, \alpha, \vecy_j) \\
& \propto \! p( \! z_j \! = \! k | \vecz_{\neg j}, \! \alpha) p ( \vecy_j | z_j \! = \! k, \! \boldsymbol{\phi}_k \!).
\end{align*}
The prior probability $p(z_j = k | \vecz_{\neg j}, \alpha)$ is given by \fref{eq:crp}. For non-empty clusters, the observed data likelihood $p ( \vecy_j | z_j = k, \boldsymbol{\phi}_k)$ is given by \fref{eq:lik}. However, this does not apply to new clusters that are previously empty. For a new cluster, we marginalize out $\boldsymbol{\phi}_k$, resulting in
\begin{align*}
p(\vecy_j | z_j = k, \beta) & = \int_{\boldsymbol{\phi}_k} p(\vecy_j | z_j = k, \boldsymbol{\phi}_k) p(\boldsymbol{\phi}_k | \beta) \\
& = \int_{\boldsymbol{\phi}_k} \textit{Mult}(\vecy_j | z_j = k,\boldsymbol{\phi}_k ) \textit{Dir}(\boldsymbol{\phi}_k | \beta) \\
& = \frac{\Gamma(V \beta)}{\Gamma( \sum_{i=1}^V Y_{i,j} + V \beta)} \prod_{i=1}^V \frac{\Gamma(Y_{i,j} + \beta)}{\Gamma(\beta)},
\end{align*}
where $\Gamma(\cdot)$ is the Gamma function. 

If a cluster becomes empty after we remove a solution from its current cluster, then  we remove it from our sampling process and erase its corresponding multinomial parameter vector $\boldsymbol{\phi}_k$. 
If a new cluster is sampled for $z_j$, then we sample its multinomial parameter vector $\boldsymbol{\phi}_k$ immediately according to Step~2 below. Otherwise, we do not change $\boldsymbol{\phi}_k$ until we have finished sampling $\vecz$ for all solutions. 

\item[2.] {\bf Sample $\boldsymbol{\Phi}$}: For each cluster $k$, sample $\boldsymbol{\phi}_k$ from its posterior $\textit{Dir}(\boldsymbol{\phi}_k | n_{1,k} + \beta, \ldots, n_{V,k} + \beta)$, where $n_{i,k}$ is the number of times feature $i$ occurs in the solutions that belong to cluster $k$. 

\item[3.] {\bf Sample $\alpha$}: Sample $\alpha$ using the approach described in \cite{westparam}.

\item[4.] {\bf Update $\beta$}: Update $\beta$ using the fixed-point procedure described in \cite{minkadirichlet}.
\end{itemize}

%Since in steps~3 and 4 we estimate the hyperparameters $\alpha$ and $\beta$, our algorithm is tuning parameter-free. 
%\subsection{Post-Processing}
%\label{sec:pp}

\sloppy
The output of the Gibbs sampler is a series of samples that correspond to the approximate posterior distribution of the various parameters of interest.  To make meaningful inference for these parameters (such as the posterior mean of a parameter), it is important to appropriately post-process these samples.  For our estimate of the true number of clusters, $\widehat{K}$, we simply take the mode of the posterior distribution on the number of clusters $K$. We use only iterations with $K = \widehat{K}$ to estimate the posterior statistics \cite{nonparamsparfab}.

In mixture models, the issue of ``label-switching'' can cause a model to be unidentifiable, because the cluster labels can be arbitrarily permuted without affecting the data likelihood. In order to overcome this issue, we use an approach reported in \cite{nonparamsparfab}. First, we compute the likelihood of the observed data in each iteration as $p(\bY | \boldsymbol{\Phi}^\ell, \vecz^\ell)$, where $\boldsymbol{\Phi}^\ell$ and $\vecz^\ell$ represent the samples of these variables at the $\ell^\text{th}$ iteration. After the algorithm terminates, we search for the iteration $\ell_\text{max}$ with the largest data likelihood and then permute the labels $\vecz^\ell$ in the other iterations to best match $\boldsymbol{\Phi}^{\ell}$ with $\boldsymbol{\Phi}^{\ell_\text{max}}$. 
We use $\widehat{\boldsymbol{\Phi}}$ (with columns $\hat{\boldsymbol{\phi}}_k$) to denote the estimate of $\boldsymbol{\Phi}$, which is simply the posterior mean of $\boldsymbol{\Phi}$.
Each solution $j$ is assigned to the cluster indexed by the mode of the samples from the posterior of $z_j$, denoted by $\hat{z}_j$.

\subsection{Auto-Grading via $\samsb$}
\label{sec:auto}

We now detail how to use $\samsb$ to automatically grade a large number $N$ of learners' solutions to a mathematical question, using a small number $\widehat{K}$ of instructor graded solutions. 
First, as in $\samss$, we select the set $\mathcal{I}_B$ of ``typical solutions'' for the instructor to grade. 
We construct $\mathcal{I}_B$ by selecting one solution from each of the $\widehat{K}$ clusters that is most representative of the solutions in that cluster:
%We construct the index set $\mathcal{I}_B$ of solutions that the instructor needs to grade as 
\begin{align*}
\mathcal{I}_B = \{ \arg \max_j p(\vecy_j | \hat{\boldsymbol{\phi}}_k), k = 1,2,\ldots,\widehat{K} \}.
\end{align*}
In words, for each cluster, we select the solution with the largest likelihood of being in that cluster.

The instructor grades the $\widehat{K}$ solutions in $\mathcal{I}_B$ to form the set of instructor grades $\{ g_k \}$ for $k \in \mathcal{I}_B$. 
Using these grades, we assign grades to the other solutions $j \notin \mathcal{I}_B$ according to
\begin{align} \label{eq:grade}
\hat{g}_j = \frac{\sum_{k=1}^{\widehat{K}} p(\vecy_j | \hat{\boldsymbol{\phi}}_k) g_k}{\sum_{k=1}^{\widehat{K}} p(\vecy_j | \hat{\boldsymbol{\phi}}_k)}.
\end{align}
That is, we grade each solution not in $\mathcal{I}_B$ as the average of the instructor grades weighted by the likelihood that the solution belongs to cluster. 
We demonstrate the performance of auto-grading via $\samsb$ in the experimental results section below.

\subsection{Feedback Generation via $\samsb$}
\label{sec:alert}

In addition to grading solutions, $\samsb$ can automatically provide useful feedback to learners on where they made errors in their solutions.  

For a particular solution $j$ denoted by its column feature value vector $\vecy_j$ with $V_j$ total expressions, let $\vecy_j^{(v)}$ denote the feature value vector that corresponds to the first $v$ expressions of this solution, with $v = \{1,2,\ldots,V_j\}$. 
%{\bf rich: do we want $K$ or $\widehat{K}$ below?}
Under this notation, we evaluate the probability that the first $v$ expressions of solution $j$ belong to each of the $\widehat{K}$ clusters: $p(\vecy_j^{(v)} | \hat{\boldsymbol{\phi}}_k), k = \{1,2,\ldots,\widehat{K}\}$, for all $v$.  
Using these probabilities, we can also compute the expected credit of solution $j$ after the first $v$ expressions via
\begin{align} \label{eq:expgrade}
\hat{g}_j^{(v)} = \frac{\sum_{k=1}^{\widehat{K}} p(\vecy_j^{(v)} | \hat{\boldsymbol{\phi}}_k) g_k}{\sum_{k=1}^{\widehat{K}} p(\vecy_j^{(v)} | \hat{\boldsymbol{\phi}}_k)},
\end{align}
where $\{g_k\}$ is the set of instructor grades as defined above. 

Using these quantities, it is possible to identify that the learner has likely made an error at the $v^\text{th}$ expression if it is most likely to belong to a cluster with credit $g_k$ less than the full credit or, alternatively, if the expected credit $\hat{g}_j^{(v)}$ is less than the full credit. 

The ability to automatically locate {\em where} an error has been made in a particular incorrect solution provides many benefits.  For instance, $\samsb$ can inform instructors of the most common locations of learner errors to help guide their instruction.  It can also enable an automated tutoring system to generate feedback to a learner as they make an error in the early steps of a solution, before it propagates to later steps.  We demonstrate the efficacy of $\samsb$ to automatically locate learner errors using real-world educational data in the experiments section below.  

\section{Experiments}
\label{sec:sims}

In this section, we demonstrate how $\samss$ and $\samsb$ can be used to accurately estimate the grades of roughly $100$ open response solutions to mathematical questions by only asking the course instructor to grade approximately $10$ solutions.  We also demonstrate how $\samsb$ can be used to automatically provide feedback to learners on the locations of errors in their solutions. 

\subsection{Auto-Grading via $\samss$ and $\samsb$}

\paragraph{Datasets}
Our dataset that consists of $116$ learners solving $4$ open response mathematical questions in an edX course.  The set of questions includes $2$ high-school level mathematical questions and $2$ college-level signal processing questions (details about the questions can be found in \fref{tbl:dataset}, and the question statements are given in the Appendix).  For each question, we pre-process the solutions to filter out the blank solutions and extract features.   Using the features, we represent the solutions by the matrix $\bY$ in \fref{eq:y}.  
Every solution was graded by the course instructor with one of the scores in the set $\{0,1,2,3\}$, with a full credit of $3$. 

\begin{table}
\vspace{-0.0cm}
\centering
\caption{Datasets consisting of the solutions of $116$ learners to $4$ mathematical questions on algebra and signal processing.  See the Appendix for the question statements.} 
\vspace{-0.0cm}
\label{tbl:dataset}
\scalebox{.77}{
    \begin{tabular}{ccc}
\toprule %[0.1em]
   &  No.of solutions $N$ & No.of features (unique expressions) $V$ \\ 
\midrule %[0.1em]
Question~1 & 108 & 78  \\
Question~2 & 113 & 53   \\
Question~3 & 90 & 100   \\
Question~4 & 110 & 45   \\
\bottomrule %[0.1em]
\end{tabular}}
\vspace{-0.0cm}
\end{table}

\paragraph{Baseline: Random sub-sampling}
We compare the auto-grading performance of $\samss$ and $\samsb$ against a baseline method that does not group the solutions into clusters.  In this method, we randomly sub-sample all solutions to form a small set of solutions for the instructor to grade.  Then, each ungraded solution is simply assigned the grade of the solution in the set of instructor-graded solutions that is most similar to it as defined by $\bS$ in \fref{eq:similarity}.  
Since this small set is picked randomly, we run the baseline method $10$ times and report the best performance.\footnote{Other baseline methods, such as the linear regression-based method used in the edX essay grading system \cite{ease}, are not listed, because they did not perform as well as random sub-sampling in our experiments.}

\paragraph{Experimental setup}

%\sloppy
For each question, we apply four different methods for auto-grading:

\begin{itemize}

\item Random sub-sampling (RS) with the number of clusters $K \in \{ 5, 6, \dots, 40\}$.

\item $\samss$ with spectral clustering (SC) with $K \in \{ 5, 6, \dots, 40\}$.  

\item$\samss$ with affinity propagation (AP) clustering.  This algorithm does not require $K$ as an input. 

\item $\samsb$ with hyperparameters set to the non-informative values $\alpha_\alpha = \alpha_\beta = 1$ and running the Gibbs sampling algorithm for 10,000 iterations with 2,000 burn-in iterations. 

\end{itemize}
$\samss$ with AP and $\samsb$ both automatically estimate the number of clusters $K$.  
Once the clusters are selected, we assign one solution from each cluster to be graded by the instructor using the methods described in earlier sections.

\paragraph{Performance metric}

We use mean absolute error (MAE), which measures the ``average absolute error per auto-graded solution''
\begin{align*}
\text{MAE} = \frac{\sum_{j=1}^{N-K} |\hat{g}_j - g_j |}{N-K},
\end{align*}
\sloppy
as our performance metric.
Here, $N-K$ equals the number of solutions that are auto-graded, and $\hat{g}_j$ and $g_j$ represent the estimated grade (for $\samsb$, the estimated grades are rounded to integers) and the actual instructor grades for the auto-graded solutions, respectively. 

\paragraph{Results and discussion}

\begin{figure}[t]
\vspace{-0.1cm}
\centering
\hspace{-.7cm}
\begin{subfigure}[t]{0.4\linewidth}
\includegraphics[scale=0.22]{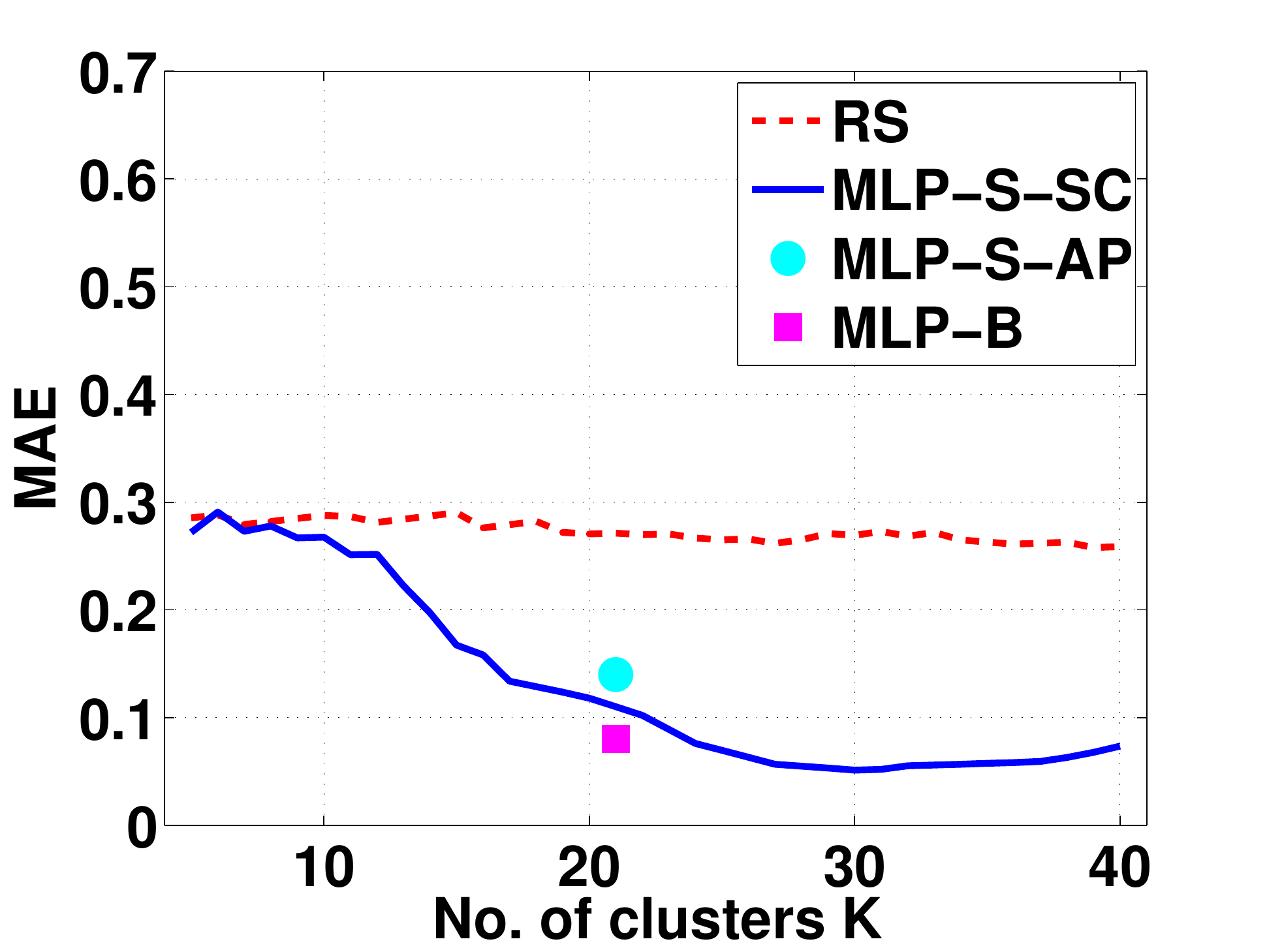}
\subcaption{Question~1}
\end{subfigure}
\hspace{.7cm}
\begin{subfigure}[t]{0.4\linewidth}
\includegraphics[scale=0.22]{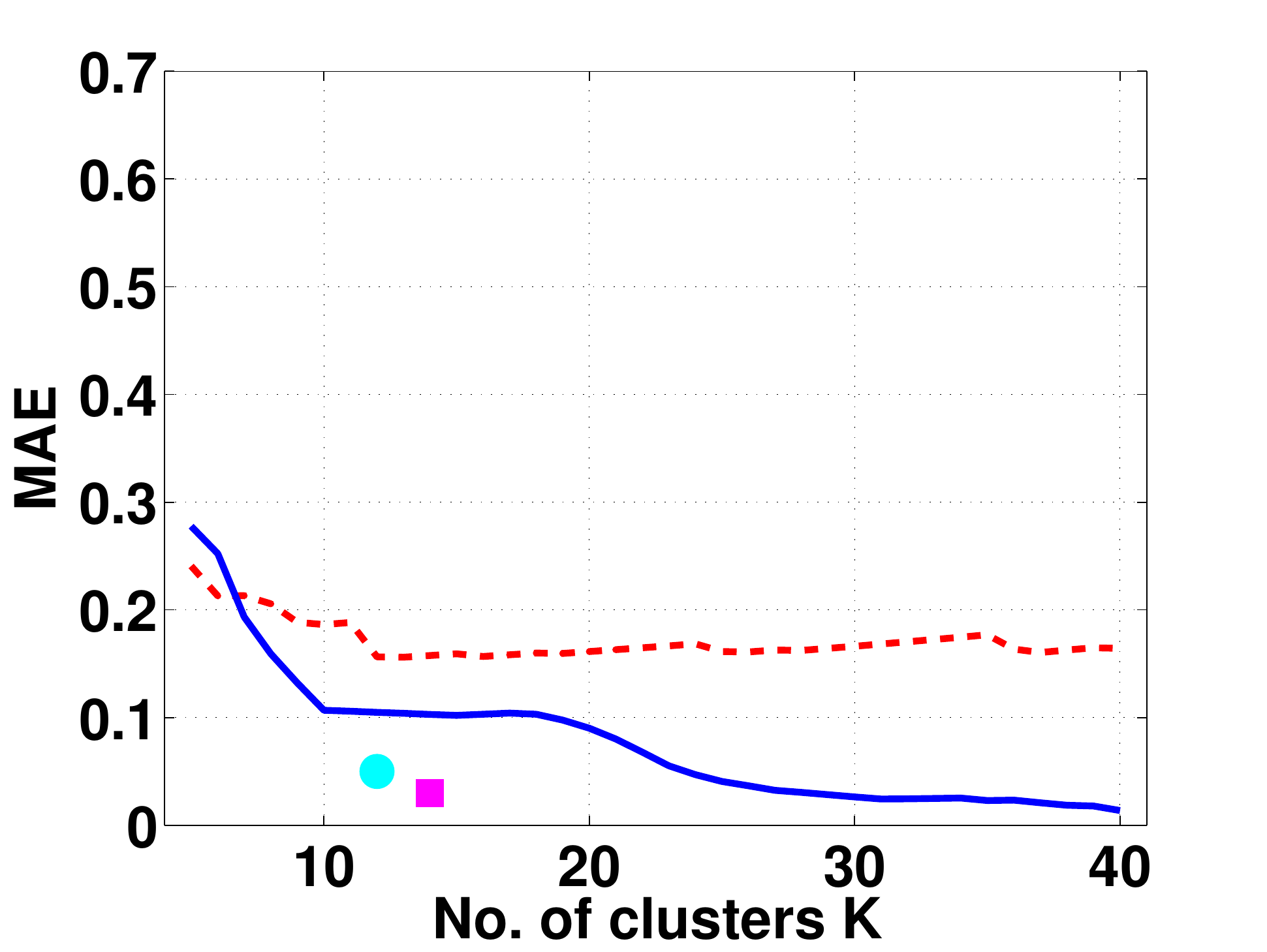}
\subcaption{Question~2}
\end{subfigure}
\\
\hspace{-.7cm}
\begin{subfigure}[t]{0.4\linewidth}
\includegraphics[scale=0.22]{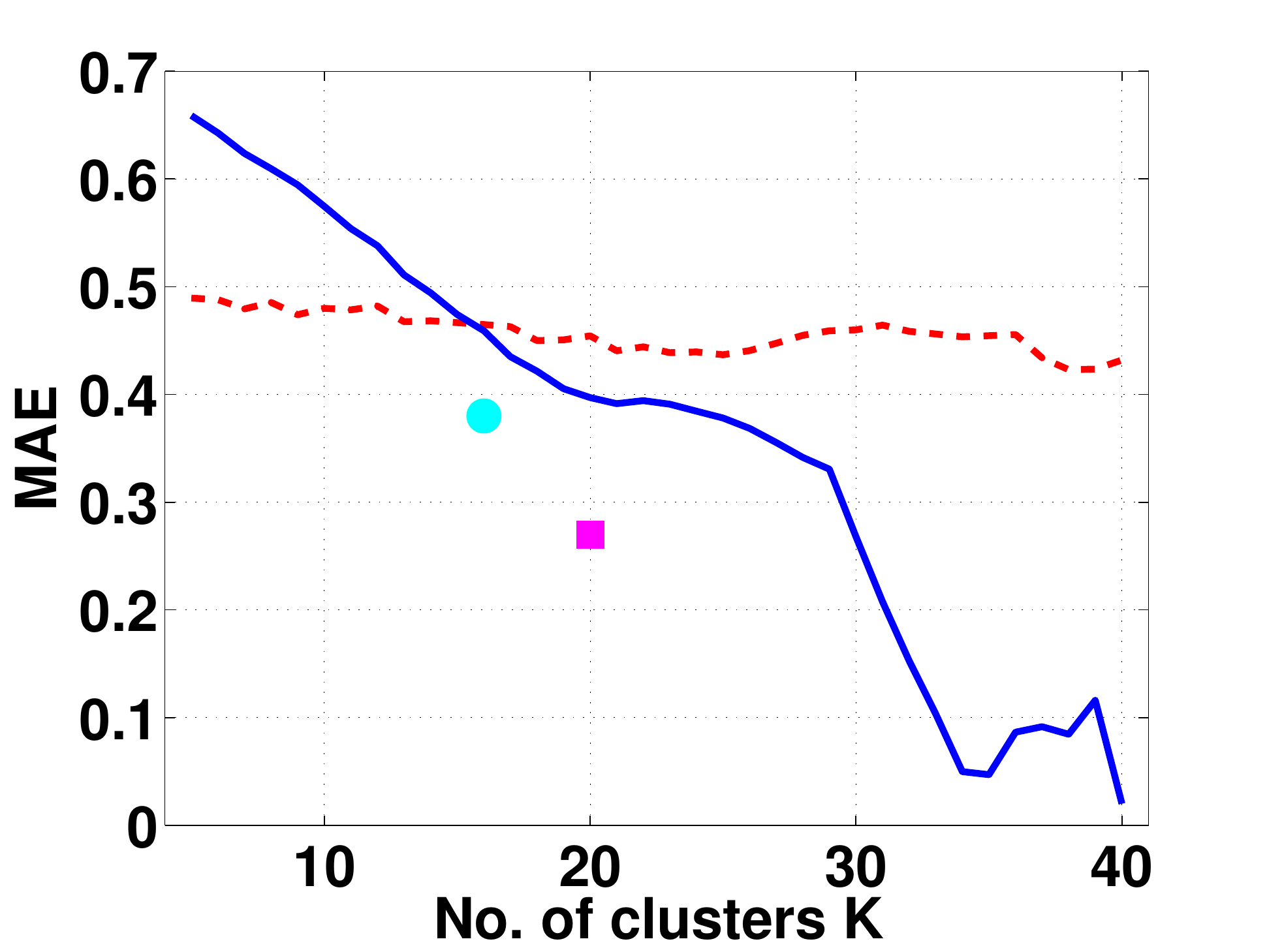}
\subcaption{Question~3}
\end{subfigure}
\hspace{.7cm}
\begin{subfigure}[t]{0.4\linewidth}
\includegraphics[scale=0.22]{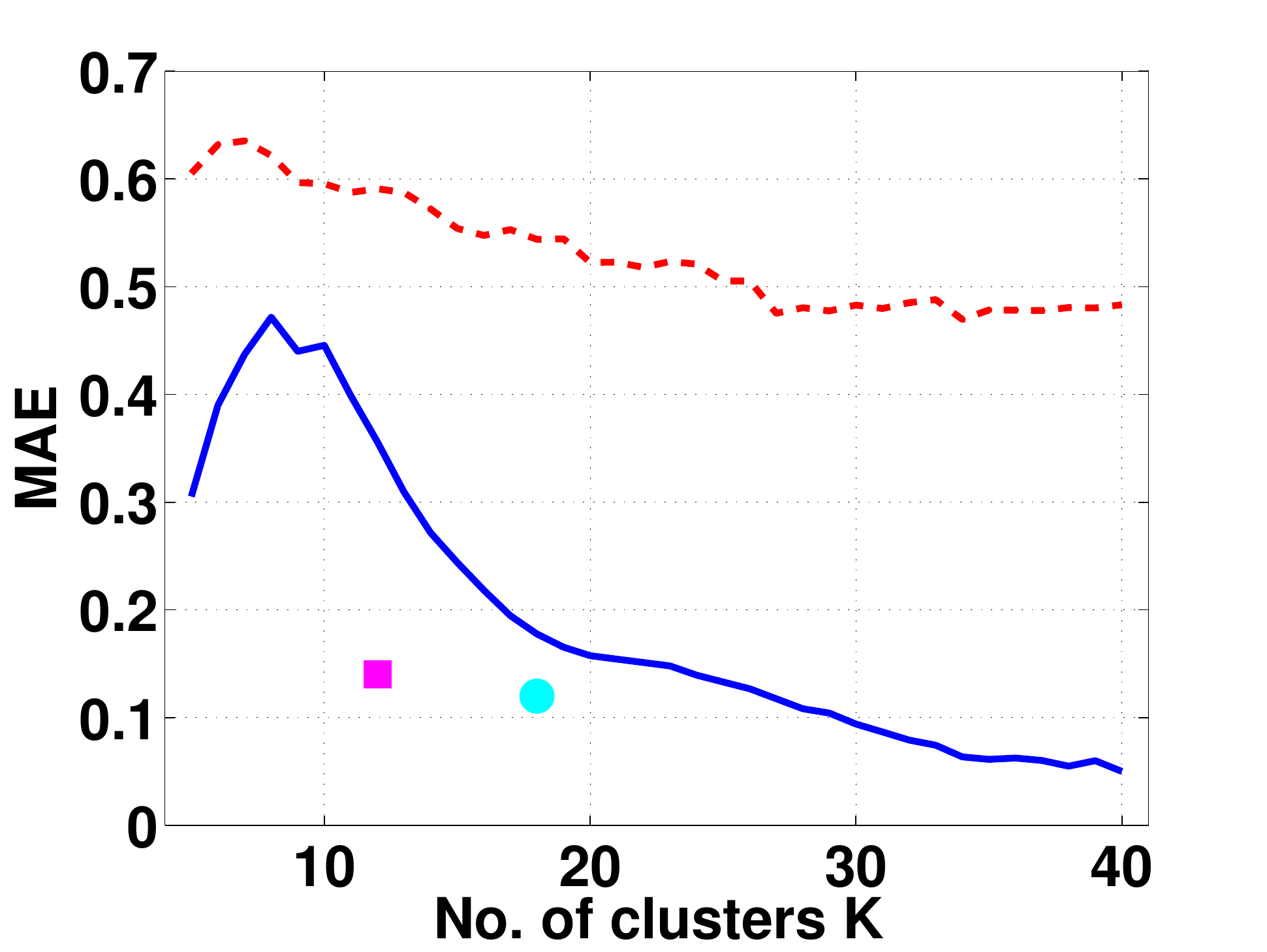}
\subcaption{Question~4}
\end{subfigure}
\caption{Mean absolute error (MAE) versus the number of instructor graded solutions (clusters) $K$, for Questions~1--4, respectively.  For example, on Question~1, $\samss$ and $\samsb$ estimate the true grade of each solution with an average error of around $0.1$ out of a full credit of $3$.  ``RS'' represents the random sub-sampling baseline. Both $\samss$ methods and $\samsb$ outperforms the baseline method.}
%, ``$\samss$-SC'' represents spectral clustering in $\samss$, ``$\samss$-AP'' represents the affinity propagation clustering in $\samss$, and ``$\samsb$'' represents $\samsb$. 
\label{fig:errorvsk}
\end{figure}

In \fref{fig:errorvsk}, we plot the MAE versus the number of clusters $K$ for Questions~1--4.  $\samss$ with SC consistently outperforms the random sampling baseline algorithm for almost all values of $K$.  This performance gain is likely due to the fact that the baseline method does not cluster the solutions and thus does not select a good subset of solutions for the instructor to grade.  $\samsb$ is more accurate than $\samss$ with both SC and AP and can automatically estimate the value of $K$, although at the price of significantly higher computational complexity (e.g., clustering and auto-grading one question takes $2$ minutes for $\samsb$ compared to only $5$ seconds for $\samss$ with AP on a standard laptop computer with a 2.8GHz CPU and 8GB memory). 

Both $\samss$ and $\samsb$ grade the learners' solutions accurately (e.g., an MAE of $0.04$ out of the full grade $3$ using only $K=13$ instructor grades to auto-grade all $N=113$ solutions to Question~2).  Moreover, as we see in \fref{fig:errorvsk}, the MAE for $\samss$ decreases as $K$ increases, and eventually reaches $0$ when $K$ is large enough that only solutions that are exactly the same as each other belong to the same cluster.  In practice, one can tune the value of $K$ to achieve a balance between maximizing grading accuracy and minimizing human effort.  Such a tuning process is not necessary for $\samsb$, since it automatically estimates the value of $K$ and achieves such a balance. 

% 4. some qualitative argument that the clustering-grading process makes sense (maybe display an example like 1 typical solution from each cluster, and how did these help grade other solutions). 

\subsection{Feedback Generation via $\samsb$}

\paragraph{Experimental setup}

Since Questions~3--4 require some familiarity with signal processing, we demonstrate the efficacy of $\samsb$ in providing feedback on mathematical solutions on Questions~1--2.  
Among the solutions to each question, there are a few types of common errors that more than one learner makes.  
We take one incorrect solution out of each type and run $\samsb$ on the other solutions to estimate the parameter $\hat{\boldsymbol{\phi}}_k$ for each cluster.  
Using this information and the instructor grades $\{ g_k \}$, after each expression $v$ in a solution, we compute the probability that it belongs to a cluster $p(\vecy_j^{(v)} | \hat{\boldsymbol{\phi}}_k)$ that does not have full credit ($g_k < 3$), together with the expected credit using \fref{eq:expgrade}.  Once the expected grade is calculated to be less than full credit, we consider that an error has occurred.   
%{\bf RICHB: make sure above sentence is clear.  also do you calculate after each EXPRESSION or each LINE?  "line" is a new term that was not really defined.  e.g.: an expression could easily extend across more than one line.  confusing!}

\paragraph{Results and discussion}
Two sample feedback generation process are shown in \fref{fig:alert}.  
%We see that $\samsb$ is able to calculate the probability that the learner's solution will end up in cluster that does not have full credit, and the expected grade, as the learner solves the question expression-by-expression.  
In \fref{fig:alert}(a), we can provide feedback to the learner on their error as early as Line~2, before it carries over to later lines.  
Thus, $\samsb$ can potentially become a powerful tool to generate timely feedback to learners as they are solving mathematical questions, by analyzing the solutions it gathers from other learners. 

\begin{figure}[t]
\vspace{-0.1cm}
\centering
\scalebox{.9}{
\begin{subfigure}[t]{.9\linewidth}
\centering
\[ \begin{aligned} 
& ((x^3 + \sin x) / e^x)' \\
& = ( e^x (x^3+\sin x)' - (x^3 + \sin x) (e^x)' )/e^{2x} \\
& \mathsf{prob. incorrect} = {\color{dgreen}0.11}, \quad \mathsf{exp. grade} = \color{dgreen}3\\
& = ( e^x (2x^2+\cos x) - (x^3 + \sin x) e^x )/e^{2x} \\
& \mathsf{prob. incorrect} = {\color{orange}0.66}, \quad \mathsf{exp. grade} = \color{orange}2\\
& = ( 2x^2 + \cos x - x^3 - \sin x) / e^x \\
& \mathsf{prob. incorrect} = {\color{red}0.93}, \quad \mathsf{exp. grade} = \color{orange}2 \\
& = ( x^2 (2-x) + \cos x - \sin x) / e^x \\
& \mathsf{prob. incorrect} = {\color{red}0.99}, \quad \mathsf{exp. grade} = \color{orange}2 
\end{aligned} \]
\subcaption{A sample feedback generation process where the learner makes an error in the expression in Line~2 while attempting to solve Question~1.}
\label{fig:alert_a}
\end{subfigure} }\\
\scalebox{.9}{
\begin{subfigure}[t]{.9\linewidth}
\centering
\[ \begin{aligned}
& (x^2 + x + \sin^2 x + \cos^2 x) (2x-3) \\
& = (x^2 + x + 1) (2x-3)  \\ 
& \mathsf{prob. incorrect} = {\color{dgreen}0.09}, \quad \mathsf{exp. grade} = \color{dgreen}3\\
%& = (x^2+x+1) 2x - (x^2+x+1) 3 \\
%& \mathsf{prob. incorrect} = {\color{dgreen}0.05}, \quad \mathsf{exp. grade} = \color{dgreen}3\\
& = 4x^3 + 2x^2 + 2x - 3x^2 - 3x - 3 \\
& \mathsf{prob. incorrect} = {\color{orange}0.82}, \quad \mathsf{exp. grade} = \color{orange}2\\
& = 4x^3 - x^2 - x - 3 \\
& \mathsf{prob. incorrect} = {\color{red}0.99}, \quad \mathsf{exp. grade} = \color{orange}2\\
\end{aligned} \]
\subcaption{A sample feedback generation process where the learner makes an error in the expression in Line~3 while attempting to solve Question~2.}
\end{subfigure}}
\caption{Demonstration of real-time feedback generation by $\samsb$ while learners enter their solutions. 
After each expression, we compute both the probability that the learner's solution belongs to a cluster that does not have full credit and the learner's expected grade.  An alert is generated when the expected credit is less than full credit.} 
\label{fig:alert}
\end{figure}

\section{Conclusions}
\label{sec:conclusions}

We have developed a framework for mathematical language processing $(\sams)$ that consists of three main steps: 
({\em i}) converting each solution to an open response mathematical question into a series of numerical features; 
({\em ii}) clustering the features from several solutions to uncover the structures of correct, partially correct, and incorrect solutions; and
({\em iii}) automatically grading the remaining (potentially large number of) solutions based on their assigned cluster and one instructor-provided grade per cluster.
As our experiments have indicated, our framework can substantially reduce the human effort required for grading in large-scale courses. 
As a bonus, $\samss$ enables instructors to visualize the clusters of solutions to help them identify common errors and thus groups of learners having the same misconceptions. 
As a further bonus, $\samsb$  can track the cluster assignment of each step of a multistep solution and determine when it departs from a cluster of correct solutions, which enables us to indicate the locations of errors to learners in real time.
Improved learning outcomes should result from these innovations.  

There are several avenues for continued research.  
We are currently planning more extensive experiments on the edX platform involving tens of thousands of learners.  
We are also planning to extend the feature extraction step to take into account both the ordering of expressions and ancillary text in a solution.  
Clustering algorithms that allow a solution to belong to more than one cluster could make $\sams$ more robust to outlier solutions and further reduce the number of solutions that the instructors need to grade.  
Finally, it would be interesting to explore how the features of solutions could be used to build predictive models, as in the Rasch model \cite{rasch} or item response theory \cite{lordirt}.

\section{Appendix: Question Statements}

Question~1:
Multiply \[(x^2+x+\sin^2 x + \cos^2 x)(2x-3) \]
and simplify your answer as much as possible.

Question~2: Find the derivative of $\displaystyle{\frac{x^3 + \sin(x)}{e^x}}$ and simplify your answer as much as possible.

Question~3:
A discrete-time linear time-invariant system has the impulse response shown in the figure (omitted). Calculate $H(e^{j\omega})$, the discrete-time Fourier transform of $h[n]$. Simplify your answer as much as possible until it has no summations.

Question~4: Evaluate the following summation
\[ \sum\limits_{k=-\infty}^{\infty}\delta[n-k] \, x[k-n]. \]

\section*{Acknowledgments}

\sloppy
Thanks to Heather Seeba for administering the data collection process and Christoph Studer for discussions and insights.  Visit our website \url{www.sparfa.com}, where you can learn more about our project and purchase t-shirts and other merchandise.

\balance

\bibliographystyle{acm-sigchi}

\end{document}